\documentclass[letterpaper]{article} 
\usepackage{aaai2026}  
\usepackage{times}  
\usepackage{helvet}  
\usepackage{courier}  
\usepackage[hyphens]{url}  
\usepackage{graphicx} 
\usepackage{natbib}  
\usepackage{caption} 
\usepackage{newfloat}
\usepackage{listings}
\DeclareCaptionStyle{ruled}{labelfont=normalfont,labelsep=colon,strut=off} 
\usepackage{algorithm}
\usepackage{array}
\usepackage{booktabs}
\usepackage{subcaption}
\usepackage{amsmath}
\usepackage{algpseudocode}
\usepackage{amssymb}
\usepackage{tabularx}
\usepackage{multirow}

\algrenewcommand\algorithmiccomment[1]{\hfill\textit{// #1}}
\usepackage[table]{xcolor}
\definecolor{lightgray}{gray}{0.92}

\floatstyle{ruled}
\newfloat{listing}{tb}{lst}{}
\floatname{listing}{Listing}
\title{Bridging Synthetic and Real Routing Problems via LLM-Guided Instance Generation and Progressive Adaptation}
\author{
    Jianghan Zhu\textsuperscript{\rm 1}\thanks{Work done during internship at A*STAR.},
    Yaoxin Wu\textsuperscript{\rm 2},
    Zhuoyi Lin\textsuperscript{\rm 3}\thanks{Zhuoyi Lin is the corresponding author.},
    Zhengyuan Zhang\textsuperscript{\rm 4},
    Haiyan Yin\textsuperscript{\rm 5},
    Zhiguang Cao\textsuperscript{\rm 1},
    Senthilnath Jayavelu\textsuperscript{\rm 3,7},
    Xiaoli Li\textsuperscript{\rm 4,6}
}

\affiliations{
    \textsuperscript{\rm 1}Singapore Management University, Singapore\\
    \textsuperscript{\rm 2}Eindhoven University of Technology, Netherlands\\
    \textsuperscript{\rm 3}Institute for Infocomm Research, Agency for Science, Technology and Research (A*STAR), Singapore\\
    \textsuperscript{\rm 4}Nanyang Technological University, Singapore\\
    \textsuperscript{\rm 5}Centre for Frontier AI Research (CFAR), Agency for Science, Technology and Research (A*STAR), Singapore\\
    \textsuperscript{\rm 6}Singapore University of Technology and Design, Singapore\\
    \textsuperscript{\rm 7}National University of Singapore, Singapore\\
    \{zhuj0044, zhengyua002\}@e.ntu.edu.sg, y.wu2@tue.nl,
    \{Lin\_Zhuoyi, Yin\_Haiyan, J\_Senthilnath\}@a-star.edu.sg, zgcao@smu.edu.sg,  xiaoli\_li@sutd.edu.sg
}

\begin{document}

\maketitle

\begin{abstract}
Recent advances in Neural Combinatorial Optimization (NCO) methods have significantly improved the capability of neural solvers to handle synthetic routing instances. 
Nonetheless, existing neural solvers typically struggle to generalize effectively from synthetic, uniformly-distributed training data to real-world VRP scenarios, including widely recognized benchmark instances from TSPLib and CVRPLib. 
To bridge this generalization gap, we present Evolutionary Realistic Instance Synthesis (\textbf{EvoReal}), which leverages an evolutionary module guided by large language models (LLMs) to generate synthetic instances characterized by diverse and realistic structural patterns.
Specifically, the evolutionary module produces synthetic instances whose structural attributes statistically mimics those observed in authentic real-world instances. Subsequently, pre-trained NCO models are progressively refined, firstly aligning them with these structurally enriched synthetic distributions and then further adapting them through direct fine-tuning on actual benchmark instances.
Extensive experimental evaluations demonstrate that EvoReal markedly improves the generalization capabilities of state-of-the-art neural solvers, yielding a notable reduced performance gap compared to the optimal solutions on the TSPLib (1.05\%) and CVRPLib (2.71\%) benchmarks across a broad spectrum of problem scales.
\end{abstract}

\begin{links}
    \link{Code}{https://github.com/HenryZhu1029/EvoReal}
    \link{Extended version}{https://arxiv.org/abs/2511.10233}
\end{links}

\section{Introduction}
\label{sec:intro}
Being a longstanding NP-hard challenge, Vehicle Routing Problems (VRPs) represent a classic family of combinatorial optimization problems (COPs) \cite{BENGIO2021405,cappart2023combinatorial} widely encountered in the field of logistics \cite{CATTARUZZA201751} and public transportation \cite{konstantakopoulos2022vehicle}. 
In recent years, Neural Combinatorial Optimization (NCO) methods have demonstrated remarkable success in tackling various VRPs through deep reinforcement learning and attention-based models, achieving state-of-the-art performance on synthetic VRP instances \cite{bello2017neuralcombinatorialoptimizationreinforcement,kool2018attention,li2021learning,drakulic2023bq,ma2023learning}. 
Despite this progress, existing NCO models frequently exhibit limited generalization when transitioning from synthetic, uniformly distributed training data to real-world problem instances, such as those found in benchmarks like TSPLib \cite{reinelt1991tsplib} and CVRPLib \cite{uchoa2017new}. This distributional shift significantly restricts their practical applicability \cite{joshi2006learning,bi2022learning,DBLP:conf/ijcai/GaoS00024}.
\begin{figure*}[t]
    \centering
    \includegraphics[width=0.8\textwidth]{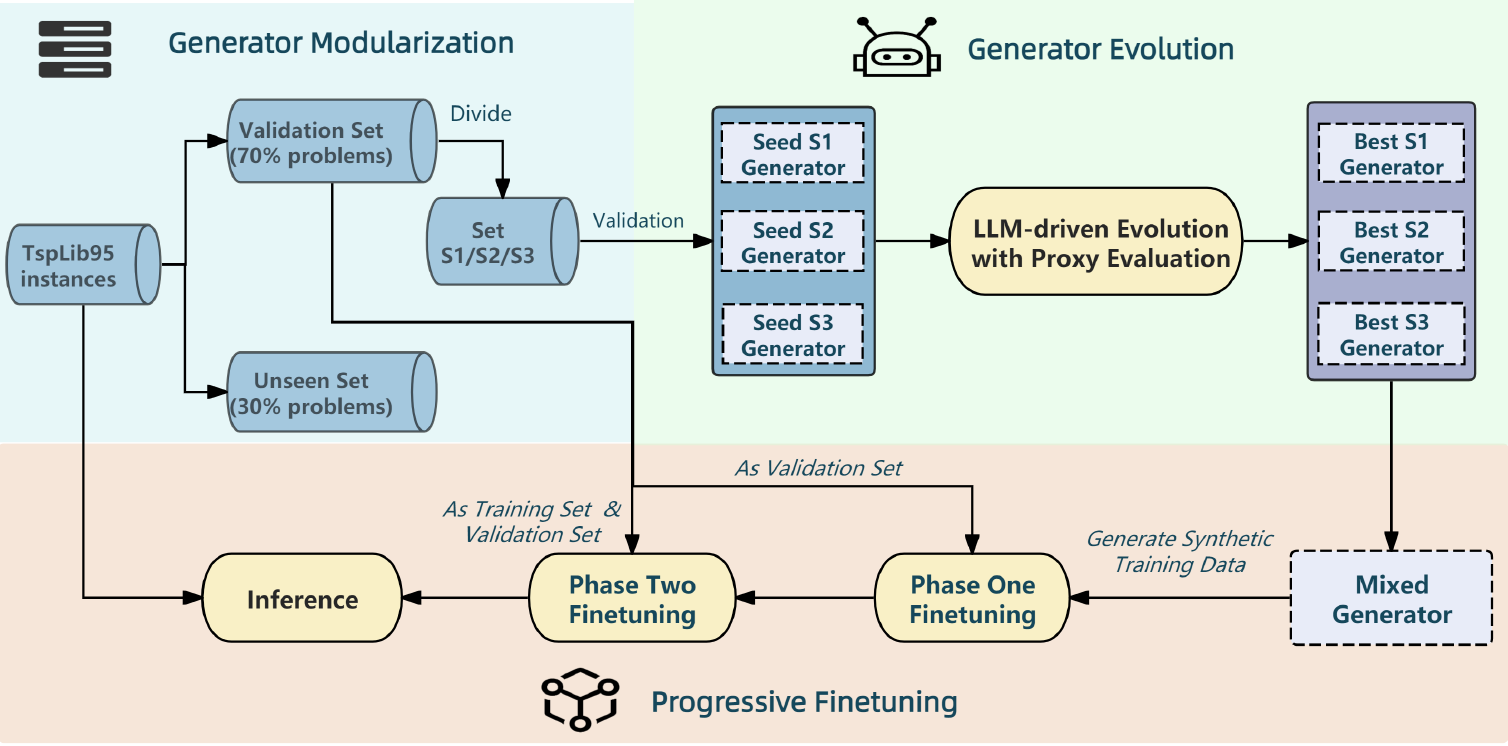}
    \caption{Overall workflow of EvoReal including LLM-guided generator evolution and progressive fine-tuning. \textbf{Top Left}: Validation set and unseen test set are split, with validation problems grouped by distribution category for structural-specific generator design. \textbf{Top Right}: LLM-driven module evolves generators which are evaluated on specific validation sets. \textbf{Bottom}: Pre-trained models are progressively fine-tuned on data from the evolved generators and the validation set’s real instances.}
    \label{fig:workflow_main}
\end{figure*}

On the other hand, LLM solvers exhibit strong generalization capabilities when confronted with new COPs and distributions.
Recent works leverage LLMs to dynamically generate heuristics or evolve hyper-heuristics \cite{duflo2019gp,drake2020recent}, thus facilitating efficient Evolutionary Search (ES) of heuristic spaces without human bias \cite{yang2024large}. Noteworthy endeavors include FunSearch \cite{romera2024mathematical}, EoH \cite{liu2024evolution}, and ReEvo \cite{ye2024reevo}. 
However, existing LLM-based approaches often encounter challenges when handling tasks that require extensive contextual information, typically demonstrating reduced coherence and accuracy in the processing of extended textual descriptions \cite{yang2024large,xu2025can}.
 Consequently, these approaches exhibit significant limitations when addressing medium- or large-size instances (e.g., instances involving more than 50 nodes), thus compromising their ability to effectively generalize to real-world VRP instances.

Motivated by these observations, we propose EvoReal, a novel data-centric framework explicitly designed to enhance the generalization capabilities of NCO models for real-world VRPs. Unlike prior approaches that evolve heuristics or solvers for direct solution construction, we innovatively leverage LLMs to evolve data generators that facilitate model adaptation.
At the core of EvoReal is an LLM-driven evolutionary module, which generates synthetic VRP instances structurally aligned with real-world distributions. This module systematically \emph{evolves simplistic synthetic scenarios into diverse and complex distributions}, effectively capturing the intricate characteristics of real-world datasets.
Subsequently, we introduce a progressive fine-tuning strategy which incrementally adapts pre-trained neural solvers by transitioning through increasingly complex synthetic instances toward actual real-world VRP scenarios.
Consequently, EvoReal \emph{facilitates smoother model adaptation by progressively refining solver parameters and representations}, effectively bridging the generalization gap between synthetic data and real-world benchmark instances. 
We summarize our contributions as follows. (1) We propose EvoReal, an LLM-guided evolutionary framework to synthesize structurally realistic VRP instances, bridging the distributional gap between synthetic and real-world routing problems. (2) We introduce a progressive fine-tuning strategy, which first adapts neural combinatorial solvers to diverse LLM-evolved distributions and then further refines them on real benchmark data, enabling effective domain adaptation without architectural changes. (3) Extensive experiments on TSPLib and CVRPLib benchmarks demonstrate that EvoReal significantly improves the generalization of SOTA neural solvers across a wide range of problem sizes, achieving new state-of-the-art results and notably reducing the performance gap between small and large instances. Notably, our results demonstrate that this generator-based adaptation consistently outperforms direct fine-tuning on real data alone.

\begin{figure*}[t]
    \centering
    \includegraphics[width=0.8\textwidth]{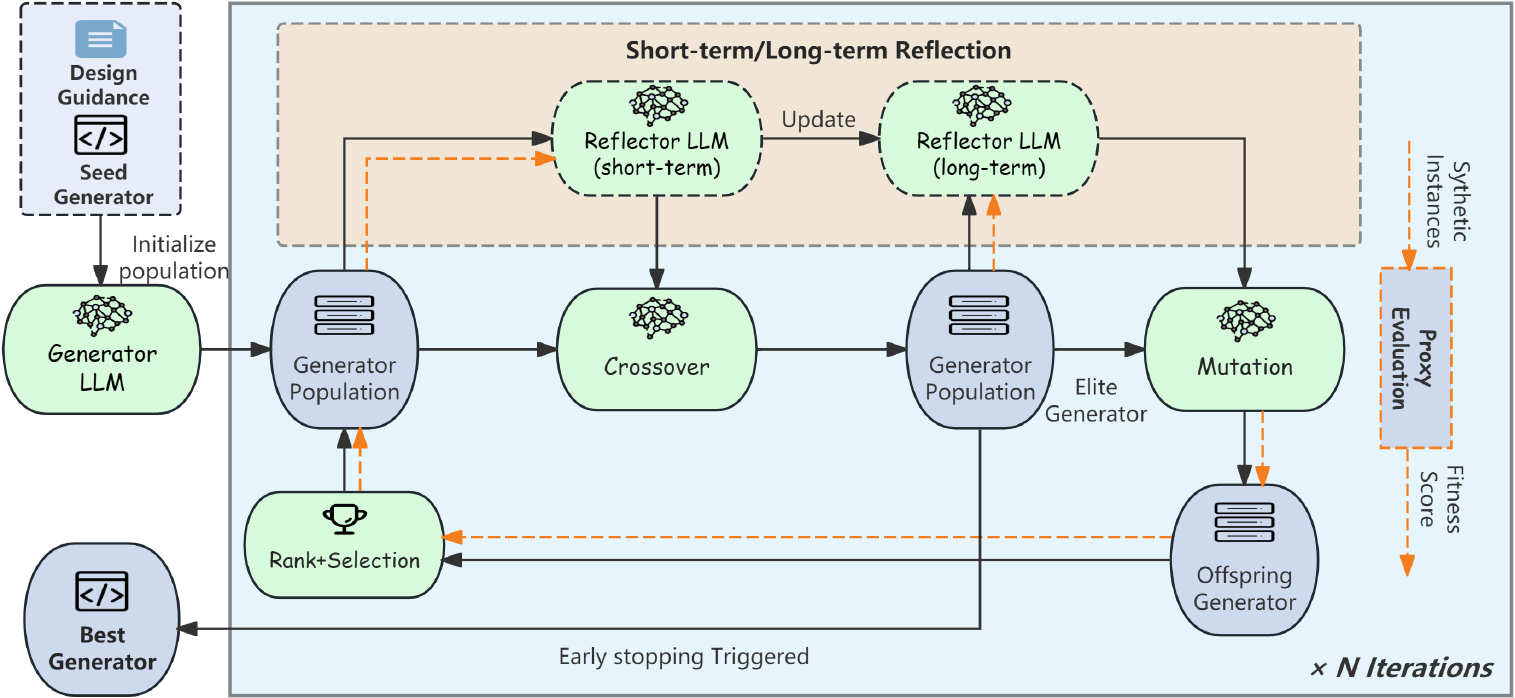}
    \caption{LLM-driven evolution component in Fig.\ref{fig:workflow_main}. The pipeline within the blue rectangle block is repeated for \textbf{$N$ }iterations. Dotted arrows represent the proxy-evaluation of each generator; black arrows indicate the flow of generators. For each pair of parents, the reflector LLM performs short-term reflection based on their relative performance, and this insight is used to guide crossover for designing new offspring. Accumulated short-term reflections are further distilled into long-term ones, which guide mutation to improve the current best generator. After mutation, populations are ranked and selected to maintain a fixed size.}
    \label{fig:workflow_evo}
\end{figure*}

\section{Related Works}
\label{sec:related_works}
\paragraph{Constructive Neural VRP Solvers.}
Those models learn policies to construct solutions in a step-by-step or one-shot manner: Pointer Network (Ptr-Net) proposed by \cite{NIPS2015_29921001} employed an encoder-decoder network that generates solutions to VRPs sequentially. Follow-up works utilize reinforcement learning (RL) to improve Ptr-Net's gradient update to generate better approximate solutions on TSP \cite{bello2017neuralcombinatorialoptimizationreinforcement} and CVRP \cite{NEURIPS2018_9fb4651c}. More recently, NCO solvers have shifted toward attention-based architectures, starting with the attention model (AM) \cite{kool2018attention}, which is versatile on a wide range of COPs. Policy optimization with multiple optima (POMO) \cite{NEURIPS2020_f231f210}, has further promoted the capability of the attention-based model on TSP and CVRP. Other constructive frameworks developed from AM and POMO illustrate their broader scalability \cite{NEURIPS2021_29539ed9,choo2022simulation,NEURIPS2022_0cddb777,chen2023neural,chalumeau2023combinatorial,hottung2024polynet,ijcai2024p769}.

Recent studies have investigated the severe generalization ability decrement in unseen problem sizes or distributions \cite{joshi2006learning,10188470}. In the line of size generalization, there are many attempts to generalize solvers from small instances to larger ones \cite{lisicki2020evaluating,kim2022scaleconditioned,bdeir2022attention,hou2023generalize,pmlr-v202-son23a}. Unlike approaches focused on scaling, considerable efforts have been devoted to addressing cross-distribution generalization, including works that augment training instances on multi-distribution and multitasks \cite{goh2025shield}, and works which synergetically train a backbone model with multi-distribution instances towards a generalizable solver on out-of-distribution tasks \cite{pmlr-v202-zhou23o,NEURIPS2024_dc709714}. 
\paragraph{Learning to Optimize with LLMs.}  Existing approaches to improving generalization in combinatorial optimization (CO) take advantage of the linguistic comprehension and domain-specific competence of LLMs via prompt engineering to tackle COPs. These methods can be mainly divided into two mainstreams: 1) \emph{LLMs are solution generators}. Early explorations in this line of research involve prompting LLMs to solve graph-based COPs \cite{NEURIPS2023_622afc4e}, iteratively refining current heuristics \cite{NEURIPS2024_eb9120be}, progressively improving solutions by targeting better objectives \cite{yang2024large,10611913,make6030093}, and instilling LLMs with visual modalities of problems and solutions \cite{elhenawy2024eyeballing,DBLP:journals/corr/abs-2403-01757}. Recently, training LLM for end-to-end CO has been attempted, moving beyond prompting paradigms~\cite{jianglarge}. 2) \emph{LLMs are heuristic designers}. Explainable, task-oriented heuristics in the form of codes are iteratively improved throughout automatic heuristic generation and evaluation of Evolution Algorithms (EA), the productivity of which outstrips that of human experts \cite{romera2024mathematical,liu2024evolution,ye2024reevo,tran2025large,jiang2025droc}. Among them, ReEvo \cite{ye2024reevo} achieves strong adaptability in many COPs under both white-box and black-box prompt settings. In this paper, we focus mainly on the utilization of LLMs as code designers. Distinct from prior constructive neural solvers and LLM-as-optimizer approaches, our method leverages LLMs to evolve data generators, enabling a novel, generator-based adaptation pipeline that outperforms direct fine-tuning on real data.

\section{Methodology}
\label{sec:methodology}
In this section, we propose a novel LLM-guided evolution framework (EvoReal) that evolves VRP data generators to generate VRPLib-style distributions and a progressive fine-tuning strategy that gradually shifts the generalization ability of the pre-trained neural model on synthetic uniform instances towards VRPLib. The proposed LLM-guided evolution approach is primarily comprised of two components: LLM-driven evolutionary search and proxy evaluation. Specifically, the \textbf{LLM-driven evolution} aims at gradually developing structurally specific generators, while in the \textbf{proxy evaluation}, performance of each LLM-designed generator is accessed throughout training and monitoring the progressive validation results. In the following part, by taking EvoReal for TSP as an example, we elaborate the generator evolution and progressive fine-tuning framework.
\paragraph{Generator Modularization.} Fig.\ref{fig:workflow_main} shows the entire pipeline of our framework. For TSP, the evolution of data generators is driven by various carefully-written prompt strategies crafted for three types of generator, which generate different distributions of TSP. This ensures the coverage of generated TSP distributions that emulate TSPLib-like distributions. We combine the three evolved generators to produce mixed-distribution TSP data for fine-tuning. 
To ensure a wide range of problem sizes and distributions, we select 70 TSPLib problems of size less than 5000, and then carefully pick 48 problems ($\approx 70 \%$) for validation while leaving the remaining 22 problems as a held-out unseen set. Among the 48 validation problems, we further categorize them into three types (i.e., S1, S2 and S3), each serving as a specific validation set for one type of generator. Detailed categorization methods are provided in the extended version. For CVRP, we randomly select 70\% problems from CVRPLib (SetX) for validation and direct fine-tuning, and the remaining 30\% are left as the unseen set. The specifics of the generation of CVRP are discussed in the extended version.

\subsection{Generator Evolution}
We prompt the LLMs to transform latent generator configurations into realistic routing distributions. Formally, let $h_\phi: \mathcal{Z} \rightarrow \mathcal{X}$ represent a data generator parameterized by an LLM model $\phi$, mapping a latent variable $z \sim \mathcal{Z}$ (e.g. seed generator) into a synthetic routing instance $x = h_\phi(z)$. Throughout the evolution, we iteratively minimize the divergence between the synthetic distribution of $h_\phi(\mathcal{Z})$ and the target real-world distribution $\mathcal{D}_{\text{real}}$: \begin{equation}
    \min_{\phi} \mathcal{L}_{\text{evolve}}(\phi) = D\left(\mathcal{D}_{\text{real}},\, \mathcal{D}\left(h_\phi(\mathcal{Z})\right)\right)
    \label{eq:evolve}
\end{equation}
where $D(\cdot \| \cdot)$ is a divergence measure (e.g., KL divergence, average gap). This dual-level optimization framework explicitly captures how heuristic generation indirectly optimizes the underlying combinatorial optimization problem by aligning evolved heuristics with real-world structures.

\noindent\textbf{Representation of Generators.} Each generator $h$ is described by three components:  1) a function description defines the format of input and describes valid output, 2) a code implementing generation, and 3) a fitness score. The fitness score, defined as a function $f$ of $h$ (i.e.   $f:h\rightarrow\mathbb{R}$), serves as the performance indicator of $h$ on the validation set. An example of an evolved generator that outputs TSP problems is given in Fig.7 in section E of the extended version.

\noindent\textbf{Proxy Evaluation.}
Following recent work \cite{DBLP:journals/corr/abs-2403-00827}, we use an approach that leverages external proxy metrics such as gaps or objective values as the divergence measure $\mathcal{D}$ in Eq.\ref{eq:evolve} to evaluate the quality of LLM-generated outputs, in lieu of direct human or reference-based evaluation. During the LLM-driven evolution, the proxy evaluation is instantiated by measuring the average gap on the validation set, which serves as the proxy metric. The gap is the gap between the objective value found and the best-known optimum. Each generator only fine-tunes the model for a small number of epochs (e.g. 10 epochs for LEHD), and the best average gap on the validation set of each generator during training is taken as the fitness score. Generators with lower fitness scores are retained according to a fixed population size, guiding generator selection without full fine-tuning. This low-fidelity evaluation strategy substantially accelerates the evolution search while presenting reliable indications of the proximity of the generated distributions to the real ones, as is the standard in the neural architecture search and hyperparameter optimization literature.

\noindent\textbf{Population Initialization.} Our evolutionary search framework begins with population initialization, during which a population of $N$ generators $h_1,h_2,...,h_N$ is initialized by incorporating initialization prompts. These prompts instill LLMs with the prior knowledge about the distributions that they are encouraged to explore and reproduce. Those prompts are composed of a trivial blueprint of the seed generator and a design guidance, each as a hint for generating possible structures of satisfactory distributions. 
Ablation studies in Section \ref{sec:ablation} show that external knowledge plays a significant role in improving the performance of generators.

\begin{table*}[t]
\centering
\setlength{\tabcolsep}{0.36mm}
\renewcommand{\arraystretch}{1.05}
\begin{tabular}{l|cc|cc|cc|cc|cc|c}
\specialrule{1.2pt}{0pt}{1pt}
\addlinespace[1.5pt]          
\toprule 
\textbf{Method} & \multicolumn{2}{c|}{\textbf{[0,200)}} & \multicolumn{2}{c|}{\textbf{[200,500)}} & \multicolumn{2}{c|}{\textbf{[500,1000)}} & \multicolumn{2}{c|}{\textbf{[1000,5000)}} & \multicolumn{2}{c|}{\textbf{Overall}} & \textbf{Time} \\
 & Obj & Gap & Obj & Gap & Obj & Gap & Obj & Gap & Obj & Gap & \\
\specialrule{0.8pt}{1pt}{1pt}
\#Problems & \multicolumn{2}{c|}{27} & \multicolumn{2}{c|}{15} & \multicolumn{2}{c|}{6} & \multicolumn{2}{c|}{22} & \multicolumn{2}{c|}{70} & \\
\midrule
Concorde & 30558.67 & 0.00\% & 42279.13 & 0.00\% & 29658.17 & 0.00\% & 158431.73 & 0.06\% & 73166.50 & 0.02\% & 8h 59min \\
LKH-3 & 30558.67 & 0.00\% & 42279.13 & 0.00\% & 29658.17 & 0.00\% & 158384.22 & 0.03\% & 73159.18 & 0.01\% & 2h 45min \\
ORTools & 31072.05 & 1.68\% & 43678.57 & 3.31\% & 30752.55 & 3.69\% & 165319.37 & 4.41\% & 75390.32 & 3.06\% & 14h 46min \\
\midrule
DIFUSCO (Ts=50) & 30830.64  & 0.89\% & 43336.11 & 2.50\% & 30758.49 & 3.71\% &  -& - & - & - & - \\
SGBS & 30803.14 & 0.80\% & 45187.94 & 6.88\% & 34750.47 & 17.17\% & 235035.03 & 48.44\% & 86670.34 & 18.48\% & 6h 29min \\
CNF (3) & 30867.31 & 1.01\% & 46266.05 & 9.43\% & 38119.65 & 28.53\% & 251882.05 & 59.08\% & 90284.04 & 23.42\% & 1h 18min\\
BQ (greedy) & 31270.68 & 2.33\% & 43623.61 & 3.18\% & 32122.76 & 8.31\% & 225740.66 & 42.57\% & 84614.77 & 15.67\% & 13min \\
BQ (bs16) & 30907.04 & 1.14\% & 43200.82 & 2.18\% & 31295.30 & 5.52\% & 216493.80 & 36.73\% & 82603.09 & 12.92\% & 4h 7min \\
ELG (no aug) & 31255.40 & 2.28\% & 44985.00 & 6.40\% & 32398.58 & 9.24\% & 178508.82 & 12.74\% & 78309.08 & 7.05\% & 7min \\
ELG ($\times$8 aug) & 30910.09 & 1.15\% & 43961.84 & 3.98\% & 32247.32 & 8.73\% & 176323.77 & 11.36\% & 77263.00 & 5.62\% & 39min \\
POMO (no aug) & 32318.85 & 5.76\% & 48481.48 & 14.67\% & 38104.81 & 28.48\% & 272069.99 & 71.83\% & 95375.41 & 30.38\% & - \\
POMO ($\times$8 aug) & 31973.53 & 4.63\% & 46701.53 & 10.46\% & 37461.23 & 26.31\% & 261002.25 & 64.84\% & 92654.16 & 26.66\% & 26min \\
LEHD (greedy) & 31179.01 & 2.03\% & 43441.81 & 2.75\% & 30859.32 & 4.05\% & 176181.27 & 11.27\% & 76999.66 & 5.26\% & 6min \\
LEHD (RRC-50) & 30659.51 & 0.33\% & 42558.18 & 0.66\% & 30289.89 & 2.13\% & 168581.11 & 6.47\% & 74966.04 & 2.48\% & 1h 51min \\
\midrule
\rowcolor{lightgray}
POMO (ours): no aug & 30946.76 & 1.27\% & 43192.36 & 2.16\% & 31366.48 & 5.76\% & 213754.57 & 35.00\% & 82259.28 & 12.45\% & - \\
\rowcolor{lightgray}
POMO (ours): $\times$8 aug & 30907.04 & 1.14\% & 43192.36 & 2.16\% & 31366.48 & 5.76\% & 209685.32 & 32.43\% & 81630.17 & 11.59\% & 26min \\
\rowcolor{lightgray}
LEHD (ours): greedy & 30986.49 & 1.40\% & 43116.26 & 1.98\% & 30666.54 & 3.40\% & 166839.40 & 5.37\% & 75302.53 & 2.94\% & 6min \\
\rowcolor{lightgray}
LEHD (ours): RRC-50 & \textbf{30650.34} & \textbf{0.30\%} & \textbf{42427.11} & \textbf{0.35\%} & \textbf{29874.67} & \textbf{0.73\%} & \textbf{162358.47} & \textbf{2.54\%} & \textbf{73919.96} & \textbf{1.05\%} & 1h 52min \\
\bottomrule                    
\addlinespace[1.5pt]
\specialrule{1.2pt}{1pt}{0pt}
\end{tabular}
\caption{Performance comparison on 70 TSPLIB problems. Each entry shows the range-wise average objective and optimality gap (\%). along with the total inference time in the last column. The best results of all learning-based methods are in \textbf{bold}.}
\label{tab:tsplib_full_objgap}
\end{table*}

\noindent\textbf{Evolution Process.} In order to balance the effectiveness of evolutionary search and the efficiency of fitness evaluation, we incorporate targeted modifications into ReEvo \cite{ye2024reevo} framework. Ours introduces a ranking-based selection mechanism that increases the survival rate of elite generators, thereby accelerating the search process and reducing token costs. In addition, by postponing the selection stage until after mutation, we balance exploration and exploitation, allowing more newly initialized generators to participate in crossover and mutation before selection. Table \ref{tab:ablation_mechanism} in Section \ref{sec:ablation} shows the elevated performance of the evolution framework equipped with these modifications. We perform the workflow in Fig.\ref{fig:workflow_evo} to search for the best generator by repeating the evolution part for N iterations. The whole evolution procedure can be summarized as the following steps:

\begin{itemize}
    \item Step 1. Choose a specific type of generator to evolve. Initialize the code population by requesting LLMs to devise $N$ generators that align with the initialization prompts.
    \item Step 2. Expand the generator population by four prompt strategies (i.e., crossover, mutation, short- \& long-term reflection) as exploration and modification operators. 
    \item Step 3. Each new generator is then evaluated on their structurally-specific validation set. Then it is added to the population if both the code and the fitness score are valid. 
    \item Step 4. Select $k$ offspring generators based on the rank of each generator individual. Ranks are calculated from fitness scores and transformed into the probability of being selected. The lower the fitness score, the more likely the generator individual will be selected.
    \item Step 5. Repeat Steps 2 to 4.
    \item Step 6. Stop the iteration once the number of the generators evaluated reaches pre-set maximum or when no better generator can be found for consecutive $m$ iterations.

\end{itemize}

Once the evolutions are completed for all types of generator, we save the best generators for subsequent progressive fine-tuning. Detailed explanations of prompt strategies and evolution mechanisms are listed in the extended version.

\subsection{Progressive Fine-tuning}
After generator evolution, we progressively fine-tune neural models to real-world distributions. Ablation studies in Section \ref{sec:ablation} show that directly fine-tuning pre-trained neural solvers with TSPLib leads to limited generalization capability improvement compared to our progressive fine-tuning strategies. Through comprehensive experimental results, we demonstrate that our progressive fine-tuning framework effectively bridges the generalization gap between vanilla synthetic distributions like uniform and heterogeneous real-world instances without modifying model architectures. We divide our progressive strategies into two phases.

\noindent\textbf{Phase One.} 
In the first phase of progressive fine-tuning, the best-performing generator of all types evolved collaboratively generate large-scale synthetic data of the same problem size. By training pre-trained models to the disparate node patterns in these diverse, TSPLib-style instances, neural solvers are better equipped to handle challenging patterns in real-world scenarios through preliminary exposure to similar structures. Compared to the proxy evaluation conducted during LLM-driven generator evolution, phase one involves a greater number of training epochs. Additionally, we continuously assess model performance throughout phase one using the 48 validation problems.

\noindent\textbf{Phase Two.} 
After aligning the neural model with a broader range of structural priors in phase one, we fine-tune the best-performed model saved in phase one with TSPLib instances and validate the model on themselves. This enables the model to shift from easy tasks (synthetic data) to accommodate to hard cases (real-world instances) seamlessly, forming a synthetic-real progression. By combining both phases together, our strategy not only enables models to generalize from synthetic to real distributions, but also goes beyond standard curriculum or domain adaptation by sequentially bridging both distributional and scale gaps.

\begin{table*}[t]
\centering
\setlength{\tabcolsep}{1.2mm}
\renewcommand{\arraystretch}{1.05}
\begin{tabular}{l|cc|cc|cc|cc|c}
\specialrule{1.2pt}{0pt}{1pt}
\addlinespace[1.5pt]          
\toprule  
\textbf{Method} & \multicolumn{2}{c|}{\textbf{[0,200)}} & \multicolumn{2}{c|}{\textbf{[200,500)}} & \multicolumn{2}{c|}{\textbf{[500,1002)}} & \multicolumn{2}{c|}{\textbf{Overall}} & \textbf{Time} \\
 & Obj & Gap & Obj & Gap & Obj & Gap & Obj & Gap & \\
\specialrule{0.8pt}{1pt}{1pt}
Count & \multicolumn{2}{c|}{22} & \multicolumn{2}{c|}{46} & \multicolumn{2}{c|}{32} & \multicolumn{2}{c|}{100} & \\
\midrule
HGS-CVRP & 27222.86 & 0.01\% & 53334.09 & 0.06\% & 102118.28 & 0.24\% & 63176.43 & 0.11\% & 16h 40min \\
LKH-3 & 27315.41 & 0.35\% & 53845.79 & 1.02\% & 103310.20 & 1.41\% & 63738.08 & 1.00\% & 27h 29min \\
ORTools & 27802.65 & 2.14\% & 55514.15 & 4.15\% & 106987.85 & 5.02\% & 65637.60 & 4.01\% & 8h 4min \\
\midrule
CNF (3) & 28964.95 & 6.41\% & 61244.12 & 14.90\% & 143774.47 & 41.13\% & 76624.53 & 21.42\% & 10min 34s\\
ELG (no aug) & 28921.39 & 6.25\% & 57342.41 & 7.58\% & 112081.53 & 10.02\% & 68199.75 & 8.07\% & 2m 15s \\
ELG ($\times$8 aug) & 28447.76 & 4.51\% & 56244.39 & 5.52\% & 109819.94 & 7.80\% & 66912.36 & 6.03\% & 5m 24s \\
POMO (no aug) & 29822.38 & 9.56\% & 63685.36 & 
19.48\% & 161266.20 & 58.30\% & 81862.41 & 29.72\% & - \\
POMO ($\times$8 aug) & 29057.50 & 6.75\% & 61270.77 & 14.95\% & 143794.84 & 41.15\% & 76693.95 & 21.53\% & 2m 17s \\
LEHD (greedy) & 30309.62 & 11.35\% & 58339.16 & 9.45\% & 119946.19 & 17.74\% & 71008.01 & 12.52\% & 2m 18s \\
LEHD (RRC-50) & 28556.65 & 4.91\% & 55823.30 & 4.73\% & 110298.74 & 8.27\% & 66830.32 & 5.90\% & 1h 10min \\
\midrule
\rowcolor{lightgray}
POMO (ours): no aug & 28279.00 & 3.89\% & 55940.56 & 4.95\% & 108505.76 & 6.51\% & 66401.20 & 5.22\% & - \\
\rowcolor{lightgray}
POMO (ours): $\times$8 aug & 28178.29 & 3.52\% & 55754.01 & 4.60\% & 108189.96 & 6.20\% & 66180.32 & 4.87\% & 2m 6s \\
\rowcolor{lightgray}
LEHD (ours): greedy & 28782.57 & 5.74\% & 55700.70 & 4.50\% & 106172.85 & 4.22\% & 66060.42 & 4.68\% & 2m 36s \\
\rowcolor{lightgray}
LEHD (ours): RRC50 & \textbf{27916.97} & \textbf{2.56\%} & \textbf{54682.63} & \textbf{2.59\%} & \textbf{104929.99} & \textbf{3.00\%} & \textbf{64817.21} & \textbf{2.71\%} & 1h 9min \\
\bottomrule                    
\addlinespace[1.5pt]
\specialrule{1.2pt}{1pt}{0pt}
\end{tabular}
\caption{Performance comparison over 100 SetX problems, with average objective, optimality gap (\%), and inference time. The best result of all learning-based methods under each problem size range are marked in \textbf{bold}.}
\label{tab:vrp_full_objgap}
\end{table*}

\section{Experimental Results}
\label{sec:exp_results}
In this section, we evaluate EvoReal on TSPLib95 \cite{reinelt1991tsplib} and CVRPLib setX \cite{uchoa2017new}.

\subsection{Experimental Settings}
\textbf{Baselines.} We employ four classic solvers as non-neural baselines:
Concorde \cite{applegate2006concorde}, ORTools, HGS \cite{VIDAL2022105643} and LKH-3 \cite{helsgaun2017extension}. For deep neural solvers, POMO \cite{NEURIPS2020_f231f210}, LEHD \cite{luo2023neural}, SGBS \cite{choo2022simulation}, BQ \cite{drakulic2023bq}, ELG \cite{DBLP:conf/ijcai/GaoS00024}, CNF \cite{NEURIPS2024_dc709714} and DIFUSCO \cite{sun2023difusco}. Importantly, POMO is trained using reinforcement learning (RL) with policy gradient methods, whereas LEHD employs a supervised learning (SL) scheme that leverages optimal objective values. These models sample solely uniform distributions for training. 
\begin{table}[t]
\centering
\setlength{\tabcolsep}{0.45mm}
\renewcommand{\arraystretch}{1.2}
\begin{tabular}{l|c|c|c|c}
\toprule
POMO (aug$\times$8) & \textbf{[0,200)} & \textbf{[200,1000)} & \textbf{[1000,5000)} & \textbf{Time} \\
\midrule

w/o Fine-tune & 4.63\% & 14.99\% & 64.84\% & -\\
w/o Phase 1 & 1.63\% & 12.06\% & 44.85\% & 7h 41min  \\
w/o Phase 2 & \textbf{1.10}\% & 11.14\% & 43.06\%&5h 42min \\
\textbf{Full (ours)} & 1.11\%  & \textbf{6.78\%} & \textbf{36.48\%} & 6h 37min\\
\bottomrule
\end{tabular}
\caption{Comparison of different POMO fine-tuning setups. We use problem of nodes $\le500$ with batch size 4 in TSPLib for finetuning. All experiments are trained for 600 epochs. The best result for each problem range is in \textbf{bold}. }
\label{tab:pomo_finetune}
\end{table}

\begin{table}[t]
\centering
\setlength{\tabcolsep}{0.55mm}
\renewcommand{\arraystretch}{1.2}
\begin{tabular}{l|c|c|c|c}
\toprule
 LEHD - greedy & \textbf{[0,200)} & \textbf{[200,1000)} & \textbf{[1000,5000)} & \textbf{Time} \\
\midrule

w/o Fine-tune & 2.03\% & 3.12\% & 11.27\% & - \\
w/o Phase 1 & 2.55\% & 3.24\% & 8.7\% & 31 min  \\
w/o Phase 2 & 1.97\% & \textbf{2.35\%} & 6.08\% & 35 min \\
\textbf{Full (ours)} & \textbf{1.40\%} & 2.39\% & \textbf{5.37\%} & 38.5 min \\
\bottomrule
\end{tabular}
\caption{Comparison of different LEHD fine-tuning setups. We use problem of nodes $\le500$ with batch size 4 in TSPLib for finetuning. All experiments are trained for 40 epochs. The best result for each problem range are in \textbf{bold}.}
\label{tab:lehd_finetune}
\end{table}
\noindent\textbf{Training Setups.}
Hyperparameters have two sets: one for proxy evaluation and the other for progressive fine-tuning. We choose two representative attention-based neural models for our study, namely POMO and LEHD. For both sets, we leverage the evolved generators to produce TSP or CVRP of size 100 for fine-tuning. In phase two, instances from VRPLib are expanded by a pre-set batch size as a single batch, and they are expanded only once in one epoch. For LEHD, Concorde solver and HGS are adopted to obtain the optimum of TSP and CVRP, respectively. For generator evolution, we use OpenAI model o3 to construct new generators. All evolution pipelines can be conducted on a single NVIDIA RTX 3090 Ti GPU with 24GB memory, whereas the progressive fine-tuning is performed on one NVIDIA RTX A6000 GPU with 48GB memory. The complete hyperparameters are given in section D of the extended version. 

\noindent\textbf{Metric and Inference.} 
We compare the results with different metrics with respect to different problem size intervals, including average objective values, average gaps (to the optimal average tour length), and total inference time. In order to monitor the performance of models during the proxy evaluation and progressive fine-tuning, we evaluate the models' performance at fixed epochs. At each validation point, we calculate the total average gap on the validation set as the performance metric with 8-fold augmentation for POMO or greedy mode for LEHD. The inference time of classic non-neural solvers (CPU) and learning-based models (GPU) are not directly comparable. For POMO, LEHD, SGBS, CNF(3), ELG, DIFUSCO (Ts=50) we report the performance of their pre-trained models with their default settings. For BQ, we reproduced their model with their original setups. For the classic non-neural solvers for TSP, we set a maximum time limit according to problem size $n$ ($T_{max}=1.8*n$ seconds for LKH-3 and ORTools, and $T_{max}=10$ minutes for Concorde). As for non-neural solvers for CVRP, we cite the results from Table 2 of HGS paper \cite{VIDAL2022105643}. For POMO, we provide the results with and without 8 times augmentation (×8 AUG and non aug). With respect to LEHD, the inference results under both greedy mode and RRC (with 50 iterations) are documented.

\subsection{Main Results}
The main experimental results for TSPLib and CVRPLib are detailed in Table \ref{tab:tsplib_full_objgap} and Table \ref{tab:vrp_full_objgap}. As shown in Table \ref{tab:tsplib_full_objgap}, we observed that compared to recent SOTA baselines, our method, namely POMO (ours) and LEHD (ours) both significantly reduce objective values and optimal gaps across all problem sizes. Markedly, with 50 RRC iterations, LEHD (ours) dominates all other learning-based methods and ORTools across all sizes. Even compared to LEHD (RRC-50) itself, most gap reductions are more than a half, without notably increasing the inference time. POMO (ours) also outstrips solvers like SGBS, BQ and POMO, especially on large instances, both with and without augmentation. Compared to the pre-trained POMO, the gaps also dropped by more than a half. Given that the model architectures for POMO and LEHD remain intact, the inference time is only subject to trivial variance. POMO (non-augmentation) and LEHD (greedy) exhibit the lowest and second-lowest inference costs, respectively, while the costs associated with their augmented variants remain within a manageable range.

Regarding the generalization in CVRPLib, we report the same metrics as before in Table \ref{tab:vrp_full_objgap}. Similarly to their TSP counterparts, POMO (ours) and LEHD (ours) under both non-augmented and augmented versions have witnessed consistent performance improvement, particularly in large-size instances. With 50 RRC iterations, LEHD under progressive fine-tuning has outperformed all neural solvers, and the results surpass ORTools in other problem intervals whereas it is still competitive against it with 0-200 nodes. The total average gaps of ours are comparable in all problem ranges (of the largest gap difference being 2.68\%), indicating that progressive fine-tuning has significantly narrowed the performance difference between small-size instances and large-size ones. POMO under progressive fine-tuning also demonstrates improved performance over ELG and the original POMO across all problem sizes. It is worth noting that LEHD (ours) show even smaller average gap in large-size instances compared to small-size ones, possibly due to overfitting on small-size problems in the second phase of progressive fine-tuning. Additionally, all inference times vary from those of pre-trained models at a marginal level. Table 6 in the extended version also shows that with our framework, neural solvers can also generalize well on the problems unseen during training.

In summary, our EvoReal framework has substantially enhanced the models' generalization capability on VRPLib, particularly on large problems. Ablation studies in Table \ref{tab:pomo_finetune} and Table \ref{tab:lehd_finetune} further corroborate the efficacy of our method.

\begin{figure}[t]
    \centering
    \includegraphics[width=1\columnwidth]{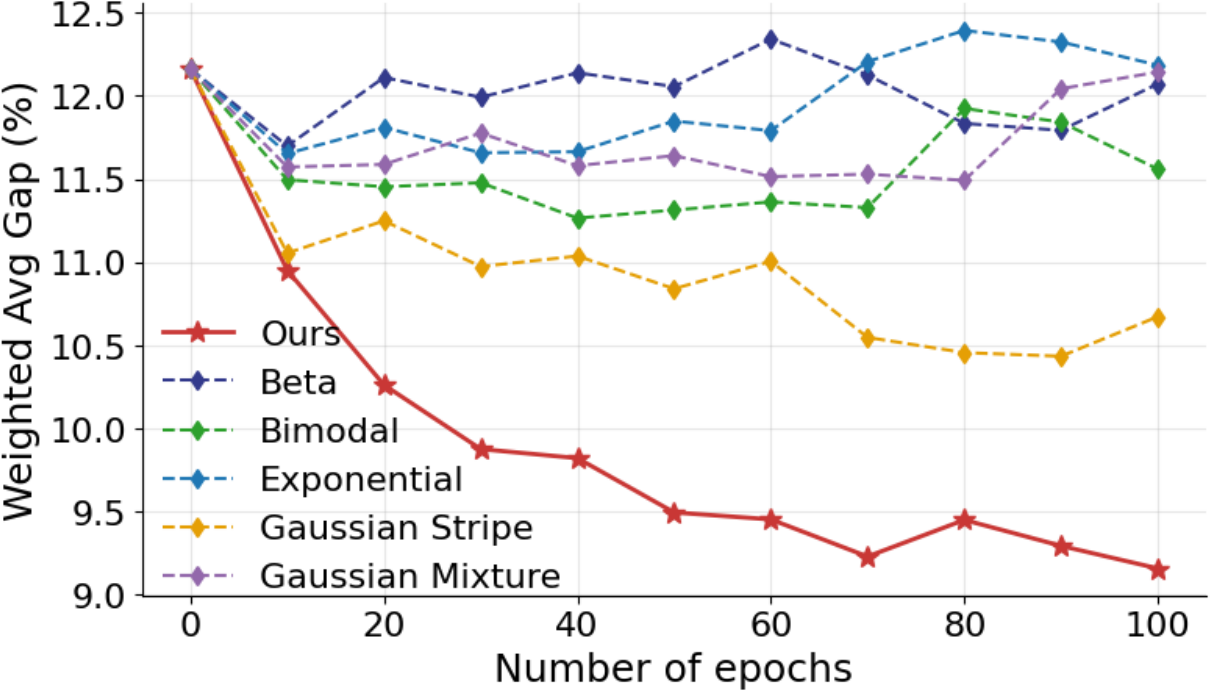}
    \caption{Comparison of the performance of the evolved generator with five the naive-distribution generators.}
    \label{fig:naive_generator}
\end{figure}

\section{Ablation Studies}
\label{sec:ablation}
\textbf{Phase Level Ablation.} We investigate the impact of each phase in progressive fine-tuning. Under two pre-trained backbone models for TSP, we ablate phase one and phase two separately, and report the range-wise average gaps and the total training time of the full progressive fine-tuning (Full), two ablation setups (w/o Phase 1 and w/o Phase 2) as well as not fine-tuning at all (w/o fine-tune). As shown in Table \ref{tab:pomo_finetune}, the gaps in our full framework consistently outperform other setups excluding problem size [0,200], in which the gap marginally degrades 0.01\% compared to without phase 2. Furthermore, compared to without phase 1, adding phase 1 markedly contributes to the performance improvement in problem size [200,1000) and [1000,5000), while the performance of without phase 1 is limited by the diversity of training instances. Similarly in Table \ref{tab:lehd_finetune}, for LEHD-greedy, our full framework is superior on both small-size and large-size problems, with only minor degradation (at 0.04\%) on the medium-size one. This further confirms that our progressive design can synergistically narrow the gap while maintaining an acceptable training cost. Fig.6 in the extended version displays the training curves corresponding to Table \ref{tab:pomo_finetune} and \ref{tab:lehd_finetune}.

\noindent\textbf{Comparison with Naive Generators.} We also compare our evolved three TSP generators, combined into a single generator, with five naive distribution generators, including beta, exponential, binomial, Gaussian mixture, and Gaussian stripe. All generators fine-tune the pre-trained POMO TSP model for 100 epochs. The trends of the total average gap in Fig.\ref{fig:naive_generator} indicate that our evolved generator with mixed distribution exceeds all naive distribution generators from the outset and ultimately converges at a much lower average gap, substantiating that our evolved generator better captures the structural characteristics of TSPLib distributions.

\noindent\textbf{LLM Mechanism-level Ablation} To verify the usefulness of the modified components in our LLM-driven evolution framework, we perform the generator evolution on S1-type generator under different setups, including the removal of rank-based selection, the removal of the design-guidance prompt, and advancing the selection before evolution. We use the manually-crafted seed generator to fine-tune the pre-trained model for w/o LLM experiment. The performance of the original model is also added for comparison (w/o fine-tuning). The average fitness score of the top five elitist generators, along with the standard deviation and the fitness score of the best generator in each setup, are reported in Table \ref{tab:ablation_mechanism}.

\begin{table}[t]
\centering
\renewcommand{\arraystretch}{1.08}
\setlength{\tabcolsep}{2pt}
\begin{tabular}{l|l}
\specialrule{1.1pt}{1pt}{1pt}
\textbf{Method} & \textbf{Avg. Aug Gap (\%)} \\ 
\specialrule{0.8pt}{1pt}{1pt}
W/O Finetuning  & 22.21 $\pm$0.00\\
W/O LLM & 19.79 {\footnotesize $\pm$0.25 (best: 19.46)} \\
W/O Rank-based Selection & 16.88 {\footnotesize $\pm$0.06 (best: 16.79)} \\
W/O Design Guidance & 16.62 {\footnotesize $\pm$0.05 (best: 16.56)} \\
W/O Late Selection & 16.55 {\footnotesize $\pm$0.07 (best: 16.50)} \\

\midrule
Full EvoReal & \textbf{16.52} {\footnotesize $\pm$0.05 (best: \textbf{16.41)}} \\
\specialrule{1.1pt}{1pt}{0pt}
\end{tabular}
\caption{Ablation of key mechanisms in our evolution framework. Each entry shows the average fitness score of top-5 generators, with standard deviation and the score of best one.}
\label{tab:ablation_mechanism}
\end{table}

\section{Conclusion and Future Works}
\label{sec:conclusion}
This work introduces a novel LLM-guided evolution framework for evolving VRP data distributions, combined with a progressive fine-tuning strategy to progressively adapt neural solvers from synthetic to diverse real-world instances. Our approach enables automated discovery and evolution of real-world-aligned VRP data generators, bridging the generalization gap for two neural solvers. Experimental results on two benchmarks demonstrate that the models trained with our evolved structural-aligned generators significantly outperform strong baselines, particularly on large-size instances, without requiring architectural changes or costly retraining. For future work, we will extend the LLM-driven evolution framework to other non-VRP tasks, such as MIS and bin packing, which involve richer structural and combinatorial constraints requiring expressive generator designs. 


\section*{Acknowledgments}

This research is supported by the National Research Foundation, Singapore, under its AI Singapore Programme (AISG Award No: AISG3-RP-2022-031, and AISG-NMLP-2024-003).

\bibliography{aaai2026}

 \clearpage

 \begin{figure*}[t]
     \centering
     \begin{subfigure}[t]{0.48\textwidth}
         \centering
         \includegraphics[width=0.98\textwidth]{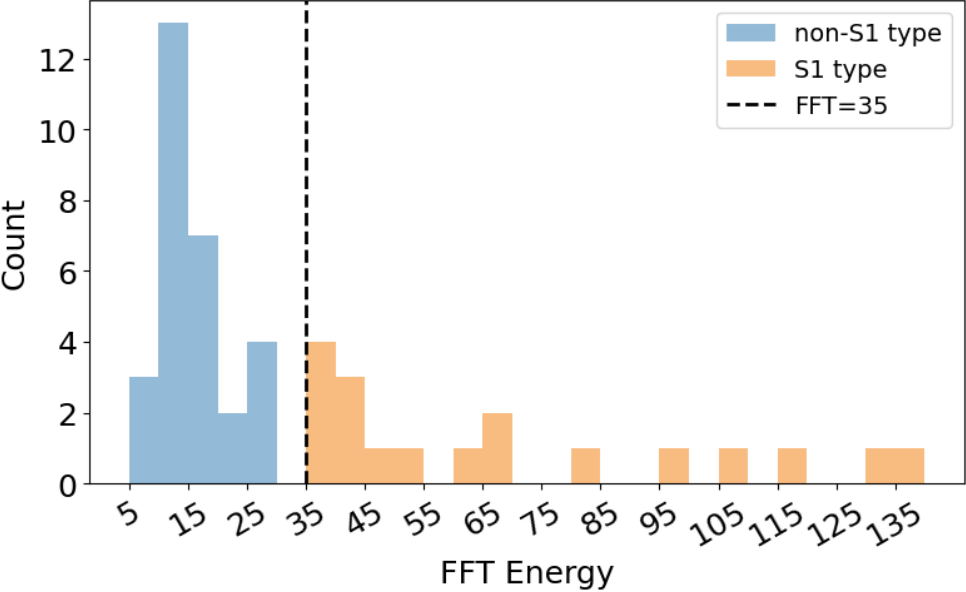}
         \caption{Histogram of the FFT energy of S1 problems and non-S1 problems in the validation set. The black dotted line represents FFT=35.}
         \label{fig:FFT_threshold}
     \end{subfigure}
     \hfill
     \begin{subfigure}[t]{0.48\textwidth}
         \centering
         \includegraphics[width=0.98\textwidth]{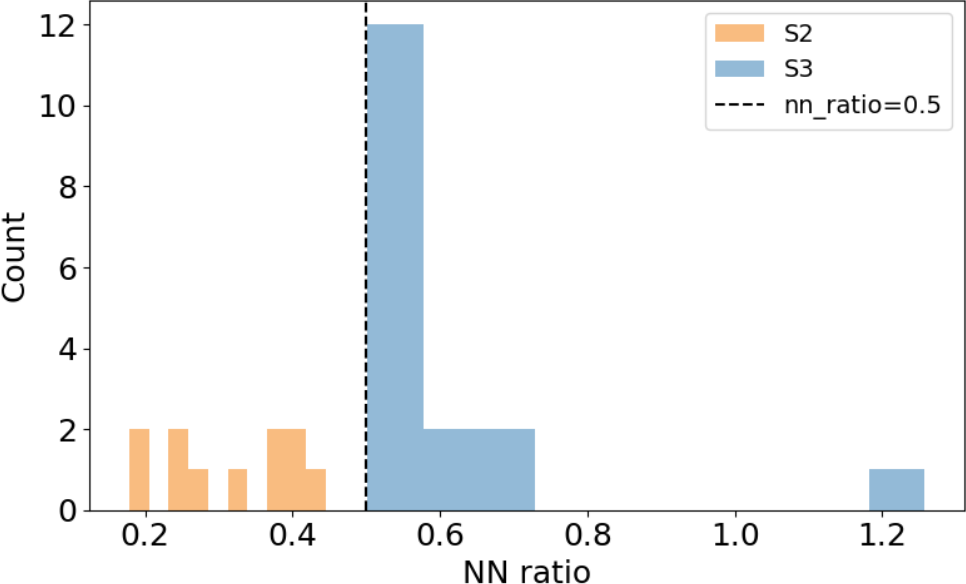}
         \caption{Histogram of the NN-ratio of S2 type instances and S3 type instances. The black dotted line represents NN ratio=0.5.}
         \label{fig:nn_ratio_threshold}
     \end{subfigure}
     \caption{Left: threshold for FFT energy of S1 type and non-S1 type. Right: threshold for NN-ratio of S2 type and S3 type. The thresholds of the divisions are both marked with black dotted line.}
     \label{fig:thresholds_fft_nn_ratio}
 \end{figure*}

 \begin{figure*}[t]
     \centering
     \begin{subfigure}[t]{0.48\textwidth}
         \centering
         \includegraphics[width=0.98\textwidth]{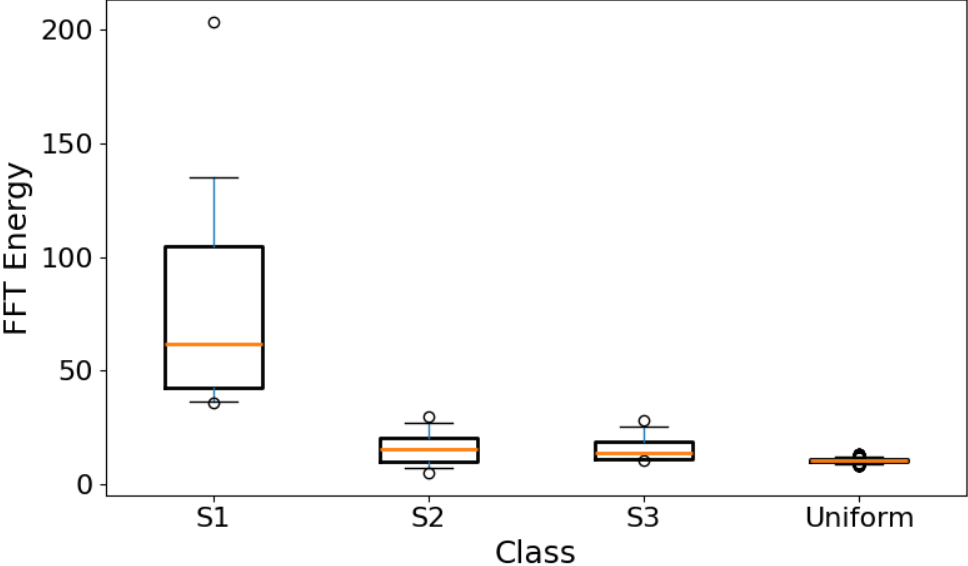}
         \caption{Box-plot for FFT energy of S1, S2 and S3 validation sets.}
         \label{fig:FFT_boxplot}
     \end{subfigure}
     \hfill
     \begin{subfigure}[t]{0.48\textwidth}
         \centering
         \includegraphics[width=0.98\textwidth]{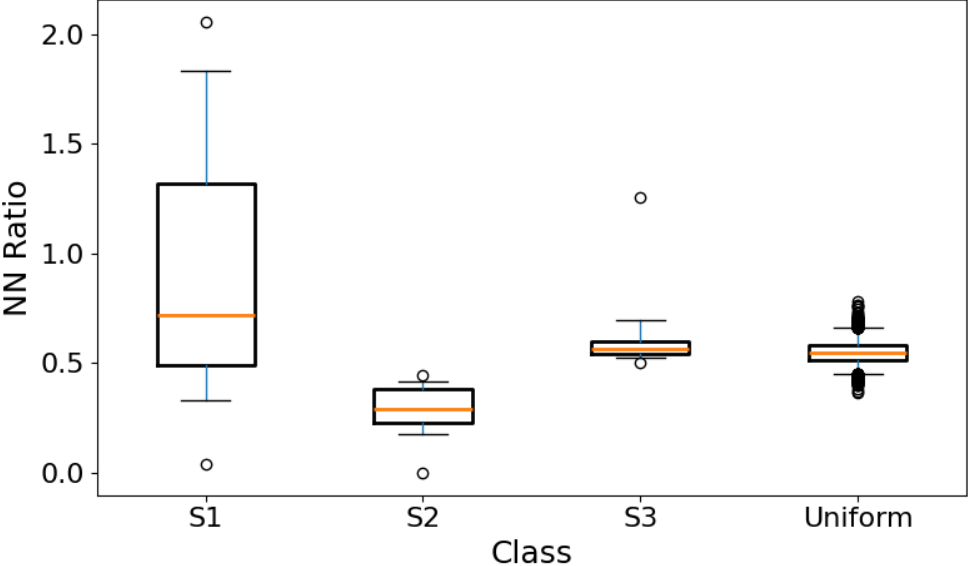}
         \caption{Box-plot for NN-ratio of S1, S2 and S3 validation problems .}
         \label{fig:nn_ratio_boxplot}
     \end{subfigure}
     \caption{Box-plots of FFT energy and NN-ratio of all validation instances after instance segmentation. The corresponding statistics of 5000 TSP 100 instances sampled from uniform distribution is supplemented for comparison.}
     \label{fig:boxplots}
 \end{figure*}

 \newpage

 \appendix

 \begin{center}
     \section*{\LARGE Supplementary Materials}
 \end{center}

 \smallskip

 \section{Pre-processing}
 \label{sec:preprocess}
 \subsection{Instance Normalization}
 We follow pre-trained models such as POMO \cite{NEURIPS2020_f231f210} and LEHD \cite{luo2023neural}, which randomly sampled uniformly distributed training instances within a unit square. To maintain consistency in data distribution and scale, we also perform the normalization procedure on the node coordinates of TSPLib \cite{reinelt1991tsplib} and CVRPLib \cite{uchoa2017new}, rescaling them to the unit square before incorporating these datasets into our training pipeline.
 The Algorithm \ref{alg:normalization} is applied to the coordinates of all nodes. We also tried other common techniques like min-max normalization,  standardization, and without-normalization for fine-tuning, but they either fail to converge or degrade the performance of the pre-trained model. We empirically find that rescaling TSPLib data to match the scale of synthetically generated uniform data, while preserving the original distributional structure, results in a more favorable convergence landscape during fine-tuning with TSPLib instances. For CVRPLib, we also employ the same approach to normalize customer nodes while re-scale the demands with capacities. For LEHD, we normalize the nodes and other constraints of the TSPLib and CVRPLib to the same scale as their training data.

 \subsection{Modularization}
 \label{sec:Categorization}
 Several prior studies have used statistics to analyze the structural diversity and heterogeneity of TSP instances \cite{herraiz2022equivalent} or predict optimal tours \cite{kou2022optimal}. Motivated by these findings, our approach adopts a structure-driven modularization strategy: we categorize TSPLib instances according to statistical features and design specialized generators tailored to each identified category. Specifically, we partition TSPLib using two complementary statistical measures: Fast Fourier Transform (FFT) energy and Nearest-Neighbor ratio (NN-ratio, also known as the Coefficient of Variation, CV). The formal definitions of these two statistics are as follows: 
 \begin{itemize}
     \item \textbf{FFT Energy:} A quantitative statistic that measures the global periodicity or repetitiveness of a spatial point pattern by evaluating the strength of frequency components on its density map. A high FFT energy indicates pronounced global regularity, periodicity, or repeated structures, while a low FFT energy suggests more disordered, chaotic structures. Inspired by classical spectral analysis in signal and spatial statistics \cite{briggs1995dft,10107467}, FFT energy is widely used to characterize spatial repetition and frequency content in spatial data.
    
     \item \textbf{NN-Ratio:} The NN-ratio is defined as the coefficient of variation of the nearest neighbor distances:
 \[
 \text{NN-ratio} = \frac{\mathrm{std}(d_{\mathrm{NN}})}{\mathrm{mean}(d_{\mathrm{NN}})}
 \]
 where \(d_{\mathrm{NN}}\) denotes all the nearest neighbor distances in the spatial point set \cite{clark1954distance,illian2008statistical}.
  A low NN-ratio indicates that most points are regularly spaced, as in grid-like or weak-clustered layouts; and a high NN-ratio indicates the presence of clusters or highly varied point spacing. It is a standard, parameter-free measure in spatial statistics to distinguish between regular, random, and clustered point patterns (with self-exclusion, that is, $k=2$ for KNN).
 \end{itemize}

 Specifically, for FFT energy, we discretize the spatial coordinates into a 2-D histogram (density map), compute the two-dimensional Fast Fourier Transform, and summarize the mean energy of the main frequency region (excluding the DC component).

 \noindent\textbf{Instance Segmentation.} We quantify the structural features of the TSPLib validation set using the two aforementioned statistics, and visualize their empirical distributions in Fig.\ref{fig:thresholds_fft_nn_ratio}. We adopt a rule-based segmentation method to classify the TSPLib instances by setting separation thresholds with these statistics. For discriminating repetitive structures, a threshold of FFT=35 is selected based on the observation that the histogram of FFT energy exhibits a clear separation at this value.: almost all non-repetitive instances (non-S1 type) have FFT energy below this threshold, while highly repetitive (S1-type) instances are distinctly partitioned above it. Similarly, to distinguish S2 from S3, a threshold of NN-ratio=0.5 is chosen, as Fig.\ref{fig:nn_ratio_threshold} shows a marked jump in the frequency distribution: all S3-type instances cluster tightly above 0.5, whereas S2-type instances are concentrated below it. The box-plots in Fig.\ref{fig:boxplots} further suggest that these thresholds form natural boundaries that maximize separability while minimizing misclassifications. The structural descriptions of the three types of distributions are as follows.
 \begin{itemize}
     \item \textbf{S1:} Instances formed by the composition of multiple repeated geometric motifs, exhibiting structured spatial regularity with controlled variation.    
     \item \textbf{S2:} Instances where points are globally arranged with low inter-nodes proximity variation, presenting weak-clustered or grid-like layouts.
     \item \textbf{S3:} Instances featured by local point aggregations, ranging from dense clusters to uniform distributions.
 \end{itemize} 
 We have also explored other unsupervised machine-learning methods for instance segmentation, such as spectral clustering \cite{von2007tutorial}, we found that the classes of most instances (85\%) remain the same as those of rule-based segmentation while resulting in trivial performance variation ($< 5 \%$) of the best evolved generators found. This consistency further substantiates the interpretability and robustness of our rule-based classification scheme, indicating that the identified structural discrepancies among different TSPLib distributions are intrinsic rather than heuristic artifacts.

 \begin{algorithm}[t]
 \caption{Normalize Coordinates}
 \label{alg:normalization}
 \begin{algorithmic}[1]
 \Require $\mathbf{X} \in \mathbb{R}^{n \times 2}$ \Comment{Input tensor with $n$ points in 2-dim}
 \Ensure $\mathbf{X}_{\text{norm}}$ \Comment{Normalized coordinates}
 \State $\text{min\_val} \gets \min(\mathbf{X}, \text{axis}=0)$
 \State $\text{max\_val} \gets \max(\mathbf{X}, \text{axis}=0)$
 \State $\text{max\_diff} \gets \max(\text{max\_val} - \text{min\_val})$
 \State $\mathbf{X}_{\text{norm}} \gets (\mathbf{X} - \text{min\_val}) / \text{max\_diff}$
 \State \Return $\mathbf{X}_{\text{norm}}$
 \end{algorithmic}
 \end{algorithm}

 \begin{figure*}[t]
     \centering
     \begin{subfigure}[t]{0.48\textwidth}
         \centering
         \includegraphics[width=0.98\textwidth]{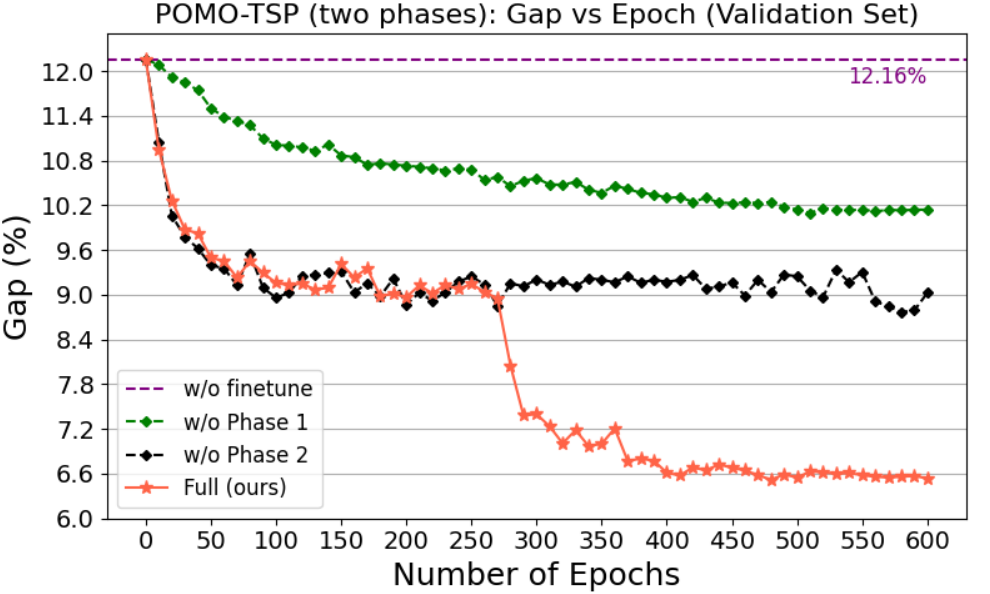}
         \caption{Training curve for POMO-TSP solver under three different ablation setups: w/o phase 1, w/o phase 2, and w/o finetune.}
         \label{fig:training_curve1}
     \end{subfigure}%
     \hfill
     \begin{subfigure}[t]{0.48\textwidth}
         \centering
         \includegraphics[width=0.98\textwidth]{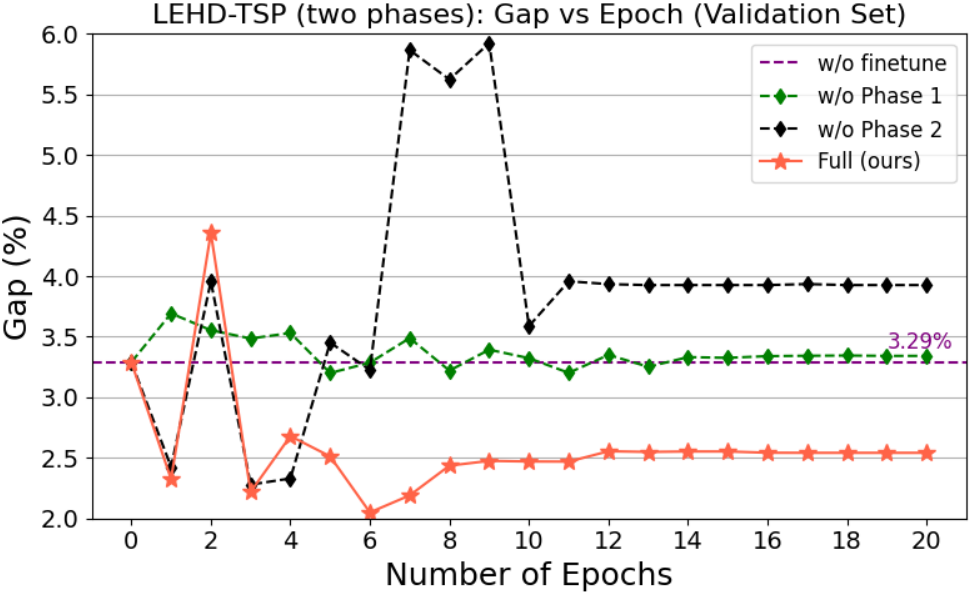}
         \caption{Training curve for LEHD-TSP solver under three different ablation setups: w/o phase 1, w/o phase 2, and w/o finetune.}
         \label{fig:training_curve2}
     \end{subfigure}
     \caption{Comparison of full progressive fine-tuning of POMO (left) and LEHD (right) TSP solver with different phase-level ablations.The ablation setups are the same as those in Table \ref{tab:pomo_finetune} and Table \ref{tab:lehd_finetune}.}
     \label{fig:training_curves}
 \end{figure*}

 \section{Evolution Mechanics}
 \label{sec:Evolution Mechanics}
 In this section, we illustrate more details regarding the LLM-guided generator evolution. Specifically, we firstly provide LLMs with detailed instructions regarding the desired distributional characteristics such as repeated or symmetric node patterns, and subsequently prompt them to design and implement corresponding generators based on these instructions. Specifically, we inform LLMs of the encouraged behaviors as well as the discouraged behaviors when designing a PyTorch-based data generator for seamless model fine-tuning. When initializing population, we formulate five categories of prompts to ensure task-aware generation:
 \begin{itemize}
     \item \textbf{Problem description:} This prompt informs LLMs about the purpose of devising a data generator.
     \item \textbf{Function format:} For formatting a valid generator, function-formatting prompts consist of function description and function signature are fabricated to standardize the names, inputs and outputs of the code.
     \item \textbf{System generator:} Further stresses on the function specifications while enumerating all safety instructions, including robust constraints and forbidden operations. 
     \item \textbf{Seed generator:} Provides LLMs with a valid exemplification of a manually designed generator for in-context learning with the structural hints and robust designs.
     \item \textbf{Design guidance:} Serves as a external knowledge constraint to encourage LLMs to explore the structures specified in the prompt.
 \end{itemize}

 \noindent\textbf{Evolution Mechanisms for Creating New Generators.}
 We build our evolution framework upon the structure of ReEvo \cite{ye2024reevo}, with modifications on several key components for better evolution-search performance. Evolution mechanisms can be classified into two groups: Searching (crossover, mutation) and Reflection (short-term reflection, long-term reflection), and they are arranged into the following sequential order in one evolution iteration:
 \begin{itemize}
     \item[] \textbf{1. Short-term reflection:} Pairs every two generators in the generator population as a parent pair, and requests LLMs to summarize the successes and failures in their design by comparing the parent generators.
     \item[] \textbf{2. Crossover:} Generates a new version of the offspring generator based on the previous parent pair along with short-term reflections.
     \item[] \textbf{3. Long-term reflection:} Synthesizes the newly gained short-term reflections and the previous long-term one.
     \item[] \textbf{4. Mutation:} Modifies the current best generator in the population under the guidance of the previous long-term reflection.

 \end{itemize}

 After mutation, the generator population is subject to probability-based selection in order to maintain a fixed population size. Unlike ReEvo, we introduced a new selection mechanism to expedite the convergence of evolution-search given the huge search space of generator structures and limited evaluation costs: 
 \begin{itemize}
     \item \textbf{Rank-based selection}: the fitness score of each generator is used to calculate the probability of survival. For example, the generators with lower best-average-augmentation-gap have a higher chance to enter the next iteration.
    
 \end{itemize}
 \noindent\textbf{Postponing early-selection.} To enhance both exploration and robustness of LLM-guided evolutionary search, we delay selection until after crossover and mutation, rather than filtering the initial population early, as in ReEvo. This prevents the premature elimination of structurally diverse or promising candidates, especially those involving complex structures, and results in a richer, more resilient generator pool for TSP data evolution. The previous ablation study has validated the superior performance of our modifications.
 \begin{table*}[t]
 \centering
 \renewcommand{\arraystretch}{1.08}
 \setlength{\tabcolsep}{1.1mm}
 \begin{tabular}{l|r|rr|rr|rr|rr}
 \specialrule{1.2pt}{0pt}{1pt}
 \addlinespace[1.5pt] 
 \toprule
 \multirow{2}{*}{Problem Name} & \multirow{2}{*}{Opt} 
 & \multicolumn{2}{c|}{POMO (×8 aug)} 
 & \multicolumn{2}{c|}{POMO (ours) (×8 aug)} 
 & \multicolumn{2}{c|}{LEHD (RRC50)} 
 & \multicolumn{2}{c}{LEHD (ours) (RRC50)} \\
 \cline{3-10}
  & & Obj & Gap & Obj & Gap & Obj & Gap & Obj & Gap \\
 \specialrule{0.8pt}{1pt}{1pt}
 berlin52 & 7542 & 9445.60 & 25.24\% & 7606.86 & 0.86\% & \textbf{7544.36} & \textbf{0.03\%} & 7544.66 & 0.04\% \\
 st70 & 675 & 699.98 & 3.70\% & 682.02 & 1.04\% & \textbf{677.11} & \textbf{0.31\%} & \textbf{677.11} & \textbf{0.31\%} \\
 pr76 & 108159 & 134906.72 & 24.73\% & 111295.61 & 2.90\% & \textbf{108159.44} & \textbf{0.00\%} & \textbf{108159.44} & \textbf{0.00\%} \\
 eil76 & 538 & 577.11 & 7.27\% & \textbf{542.41} & \textbf{0.82\%} & 544.37 & 1.18\% & 544.37 & 1.18\% \\
 pr124 & 59030 & 60192.89 & 1.97\% & 59649.82 & 1.05\% & \textbf{59030.73} & \textbf{0.00\%} & 59030.74 & 0.00\% \\
 rl1304 & 252948 & 464108.99 & 83.48\% & 333638.41 & 31.90\% & 262932.00 & 3.95\% & \textbf{256881.20} & \textbf{1.55\%} \\
 rl1323 & 270199 & 493977.81 & 82.82\% & 357013.94 & 32.13\% & 281240.97 & 4.09\% & \textbf{275050.94} & \textbf{1.80\%} \\
 nrw1379 & 56638 & 76404.66 & 34.90\% & 71375.21 & 26.02\% & 63142.54 & 11.48\% & \textbf{58333.34} & \textbf{2.99\%} \\
 fl1400 & 20127 & 28511.91 & 41.66\% & 23258.76 & 15.56\% & 21139.27 & 5.03\% & \textbf{20770.40} & \textbf{3.20\%} \\
 u1432 & 152970 & 205698.76 & 34.47\% & 183548.70 & 19.99\% & 157660.63 & 3.07\% & \textbf{154439.91} & \textbf{0.96\%} \\
 fl1577 & 22249 & 35665.15 & 60.30\% & 29947.15 & 34.60\% & 24010.49 & 7.92\% & \textbf{22680.66} & \textbf{1.94\%} \\
 d1655 & 62128 & 106487.39 & 71.40\% & 83089.99 & 33.74\% & 66893.05 & 7.67\% & \textbf{64340.05} & \textbf{3.56\%} \\
 vm1748 & 336556 & 643293.14 & 91.14\% & 439003.65 & 30.44\% & 350746.59 & 4.22\% & \textbf{341082.19} & \textbf{1.34\%} \\
 u1817 & 57201 & 113349.50 & 98.16\% & 79366.39 & 38.75\% & 60742.67 & 6.19\% & \textbf{59037.48} & \textbf{3.21\%} \\
 rl1889 & 316536 & 647569.35 & 104.58\% & 446410.72 & 41.03\% & 331794.13 & 4.82\% & \textbf{323538.38} & \textbf{2.21\%} \\
 d2103 & 80450 & 139709.47 & 73.66\% & 114769.97 & 42.66\% & \textbf{84679.73} & \textbf{5.26\%} & 85015.16 & 5.67\% \\
 u2152 & 64253 & 127041.03 & 97.72\% & 90975.82 & 41.59\% & 69816.06 & 8.66\% & \textbf{66345.79} & \textbf{3.26\%} \\
 u2319 & 234256 & 293991.28 & 25.50\% & 272791.11 & 16.45\% & 240811.14 & 2.80\% & \textbf{235586.16} & \textbf{0.57\%} \\
 pr2392 & 378032 & 674068.86 & 78.31\% & 536087.18 & 41.81\% & 415166.63 & 9.82\% & \textbf{389460.37} & \textbf{3.02\%} \\
 pcb3038 & 137694 & 234782.04 & 70.51\% & 193019.45 & 40.18\% & 151426.67 & 9.97\% & \textbf{142091.75} & \textbf{3.19\%} \\

 fl3795 & 28772 & 54635.15 & 89.89\% & 43839.90 & 52.37\% & 31647.42 & 9.99\% & \textbf{29807.47} & \textbf{3.60\%} \\
 fnl4461 & 182566 & 330827.85 & 81.21\% & 263059.35 & 44.09\% & 209771.78 & 14.90\% & \textbf{191593.55} & \textbf{4.94\%} \\

 \midrule
 Average & 128614.50 & 203597.92 & 58.30\% & 163105.40 & 26.82\% & 136344.44 & 5.52\% & \textbf{131455.05} & \textbf{2.21\%} \\
 \bottomrule
 \addlinespace[1.5pt]
 \specialrule{1.2pt}{1pt}{0pt}
 \end{tabular}
 \caption{Performance on 22 Unseen TSPLib Problems with original POMO and LEHD, and the ones trained with our EvoReal framework. The best gap and the best objective value of each problem are marked in \textbf{bold}.}
 \label{tab:tsplib_22unseen}
 \end{table*}
 \begin{table}[t]
 \centering
 \setlength{\tabcolsep}{1.5mm}
 \begin{tabular}{p{6cm}|l}
 \specialrule{1.2pt}{1pt}{1pt}
 \textbf{Parameter} & \textbf{Value} \\
 \hline
 LLM model (generator and reflector)& OpenAI: o3 \\
 Temperature & 1 \\
 Initial population size & 30 \\
 Offspring population size & 10 \\
 Crossover rate & 1 \\
 Mutation rate & 0.5 \\
 Maximum number of iterations & 10 \\
 Maximum number of evaluations & 125\\
 \midrule
 POMO (Kwon et al., 2020)& \\
 Epochs & 40\\
 Learning rate decay & No decay \\
 Train episodes per epoch & 5000 \\
 \midrule
 LEHD (Luo et al., 2023)& \\
 Epochs & 10\\
 Learning rate decay & 0.97 \\
 Train episodes per epoch& 10000 \\
 \specialrule{1.2pt}{0pt}{1pt}
 \end{tabular}
 \caption{Experiment settings of generator evolution.}
 \label{tab:experiment-settings}
 \end{table}

 \section{Routing Problem Formulation}
 \label{sec:TSP_CVRP}
 \textbf{TSP.} In TSP, each node $v_i \in V$ is characterized by a 2-D coordinate $v_i = (x_i,y_i)$. The weight of each edge between any two nodes corresponds to their Euclidean distance, computed by $w_{ij} = \sqrt{(x_i-x_j)^2+(y_i-y_j)^2}$. A solution is considered feasible if each node $v_i$ is visited exactly once in a tour and the tour is closed. The objective is to find the tour with a minimum total length. 

 \noindent \textbf{CVRP. }\cite{10.5555/2723809} Every node in CVRP has two elements: node coordinate $v_i$ and demand $\delta_i$ for $n \in \{1,2,3,...n\}$. Specifically, the demand $\delta_0$ for the depot node $v_0$ is 0. The solution to the problem consists of multiple subtours, each starting and terminating at the depot. A feasible tour should satisfy: 1) every customer node in the problem is visited only once. 2) The total demand in each subtour does not exceed the pre-defined vehicle capacity. 

 In this paper, we mainly consider the evolution of the distributions of the customer nodes, which is consistent with the design of our framework. To craft the seed generator for CVRP data evolution, we also categorize CVRP instances into three sets according to different depot positioning schemes \cite{GOUVEIA1996178}: 1) S1: depot is located in the center of the unit square. 2) S2: depot is randomly sampled from the unit square $[0,1]^2$. 3) S3: depot is at (0,0). Nonetheless, instead of evolving the generator of each category one at a time, we incorporate these three types of depots into one generator to save evolution cost. Following existing works \cite{uchoa2017new,NEURIPS2018_9fb4651c}, the coordinates of the customer nodes are sampled uniformly from the unit square while the demand $d_i$ of each customer node is initialized with $U[1,10]$. The capacity $Q$ of each vehicle is set according to the following formula: $\mathrm{Q} = \max \left(
     \left\lceil r \cdot \frac{1}{n} \sum_{i=1}^{n} d_i \right\rceil,\,
     \left\lceil k \cdot \max_{1 \leq i \leq n} d_i \right\rceil
 \right)$ where $r \sim \mathrm{Triangular}(3,\,6,\,25)$ and $k$ is a constant (e.g. $k=2$). The demands and capacity are further normalized to $\delta_{i}^{\prime} = \delta_{i}/Q$ and 1, respectively.

 \section{Experimental Setup}
 \label{sec:Experimental Setup}

 \subsection{Model Setup and Training}
 Our framework utilizes the identical network topologies of POMO \cite{NEURIPS2020_f231f210} and LEHD \cite{luo2023neural}. For proxy evaluation in LLM-guided evolution and phase one of progressive fine-tuning, we employed the pre-trained TSP and CVRP models trained with problems of size 100, using their released checkpoints. For proxy evaluation and phase one fine-tuning, we also generate TSP100 and CVRP100 for training. Additionally, we employ the same data loader as POMO's and LEHD's. Specifically, for fine-tuning LEHD in phase one, training instances are generated only once and shuffled for reuse when training every epoch. For the results in Table \ref{tab:tsplib_full_objgap} and Table \ref{tab:vrp_full_objgap}, we set the hyperparameters in Table \ref{tab:experiment-settings} for LLM-guided evolution and proxy evaluation. 

 \begin{figure}[t]
     \centering
     \includegraphics[width=1\columnwidth]{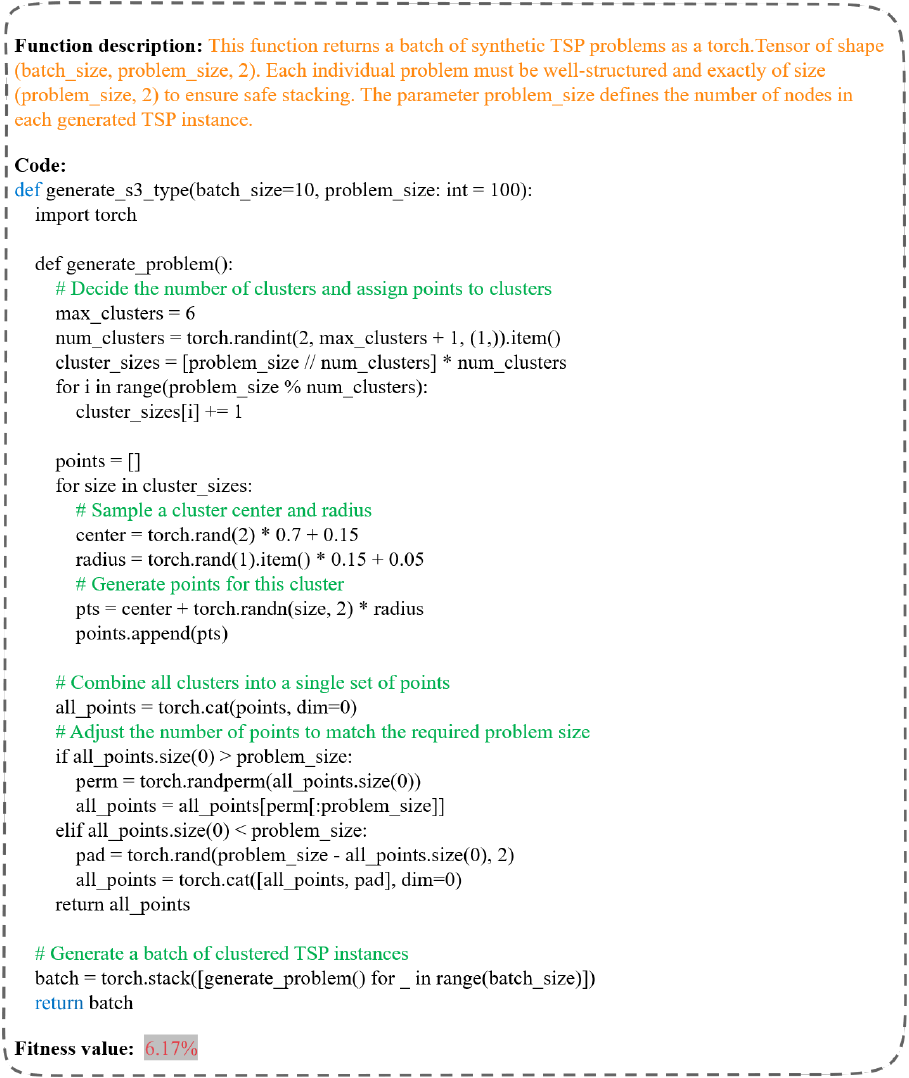}
     \caption{An example of an evolved S3-type generator that outputs TSP problems of 100 node, along with function description and its fitness score.}
     \label{fig:code_demo}
 \end{figure}
 \subsection{Hyperparameter Selection}
 \textbf{Proxy-evaluation \& Phase One Fine-tuning Setup.}
 For proxy-evaluation in LLM-based evolution and phase one fine-tuning, we maintain most of the hyperparameters the same as the original experimental settings in the respective model papers: Adam \cite{kingma2014adam} optimizer is used with an initial learning rate of $10^{-4}$ and a decay of 0.9 for both TSP and CVRP, except for the LEHD TSP model, where a lower learning rate of $10^{-5}$ and a decay of 0.6 is used. We use the same batch sizes as the original papers. However, for phase one fine-tuning, we fine-tuned the pre-trained models with more epochs and instances to see the convergence. For POMO in phase one, the model for TSP is fine-tuned for 300 epochs with 5000 problems in each episode, while the model for CVRP is fine-tuned for 200 epochs with 500 problems in each episode. For LEHD in phase one, we fine-tuned their pre-trained TSP version and their CVRP version both for 20 epochs. And in each episode for LEHD, we generate 50000 synthetic data once for the repetitive usage in each epoch. For finetuning both models in phase one, we generate mixed distributions of S1, S2, S3 according to the ratio of S1, S2, S3 problems in the validation set in one epoch (e.g. if S1 problems: S2 problems: S3 problems= 17:19:12, then the ratio of the synthetic instances of type S1, S2, S3 is also 17:19:12).

 \noindent\textbf{Phase Two Fine-tuning Setup.}
 In the case of phase two fine-tuning, the hyperparameter setups are more empirical and pragmatic. The training epochs for both problems of both models are identical to those of phase one. For POMO, we used the same learning rate and decay as those of phase one. Nevertheless, the learning rate and the decay for LEHD-TSP are $10^{-6}$ and 0.9 while the learning rate and the decay for LEHD-CVRP are $10^{-5}$ and 0.8. As for batch sizes, since we use the real instances from TSPLib and CVRPLib for fine-tuning, we expand each problem once by batch size as one batch in an epoch. Limited by computational resources, we used an adaptive batch size ($bs$) setting for both VRP problems for POMO (for POMO-TSP, we set $bs=4$ for $n \in[0,500)$, $bs=2$ for $n\in[500,750)$, $bs=1$ for $n \in[750,1002)$ where $n$ is the problem size; for POMO-CVRP, $bs=4$ for $n \in[0,500)$, $bs=2$ for $n \in [500,650)$, $bs=1$ for $n\in[650,800)$), which still achieve acceptable generalization performance on both the small- and large-size instances. As for LEHD, we used a fixed batch size and decay of 4 and 0.6 for TSP respectively, and fix a batch size and a decay of 16 and 0.8 for CVRP.\\

 \noindent \textbf{Training Curves for Phase-level Ablation.}
 As shown in Fig.\ref{fig:training_curves}, we compared our full progressive fine-tuning method with ablation setups on different phases with our backbone models, which corresponds to the results in Table \ref{tab:pomo_finetune} and Table \ref{tab:lehd_finetune}. These figures further corroborate that our full framework enables the optimization of model parameters toward better generalization optima, guiding the models to achieve dominant performance.

 \section{Additional Results}
 \label{sec:add_results}

 \subsection{Generalization on Unseen Data}
 To further evaluate generalization ability of the models trained with our framework, particularly on the data unseen during the generator evolution and progressive fine-tuning, we tested them directly on 22 held-out problems. The 22 unseen problems include 5 small-size instances and 17 instances of sizes n $\in$  $[1300,5000)]$ proposed in TSPLib95 \cite{reinelt1991tsplib}. Original POMO \cite{NEURIPS2020_f231f210} and LEHD \cite{luo2023neural}, along with those trained from progressive fine-tuning, are employed to solve the instances, with both models tested under their augmentation version (×8 augmentation or RRC-50). We report the optimum, the objective values, and the gaps for each problem in 
 Table \ref{tab:tsplib_22unseen}. As shown, LEHD trained from progressive finetuning (i.e., LEHD (ours) in the table) outperforms other models in most problems, especially on large-size ones like pcb3038 and fnl4461, while it is still comparable to the original model on the first five small instances. As for POMO (ours), all problems have seen reductions in gaps by $0.92\%~\sim~60\%+$. These results demonstrate that the models with our methods, despite not being exposed to these unseen TSPLib instances during fine-tuning, consistently achieve strong generalization performance, particularly on large-scale and challenging cases. This improvement can be largely attributed to the structural prior alignment introduced in the first phase of progressive fine-tuning, which enables the model to internalize diverse distributional patterns and transfer this knowledge to previously unseen problem instances.

 \subsection{Effect of different LLMs (for Generator Design)}
 We evaluate the effectiveness of the generator design using four up-to-date LLMs developed by OpenAI: GPT-4o, GPT-4.1, GPT-4.1 mini, and OpenAI o3 to evolve our S1-type generators for TSP. All experiments were conducted in identical settings to ensure a fair comparison. Our experimental results indicate that integrating the LLM-guided generator for phase one fine-tuning consistently improves the performance on the S1 validation set. Among the language models evaluated, OpenAI o3 achieves the best performance, suggesting that our framework is able to achieve better results with strong LLMs. 

 \section{Prompts}
 \label{sec:prompts}

 \begin{minipage}[t]{0.48\textwidth}
 \raggedright
 Prompts adopted for EvoReal are gathered in this section. 
 For evolution operators (crossover and mutation) and reflection operators (long-term reflection and short-term reflection), since we only made slight modifications to those of ReEvo \cite{ye2024reevo}, these prompts are not listed here.
 \end{minipage}%
 \hfill
 \begin{minipage}[t]{0.48\textwidth}
 \raggedright
 Prompts 1–3 are shared by both the TSP and CVRP versions of EvoReal. Problem-specific prompt components are prompt 4-10 as well as Table \ref{tab:gen_desc} and Table \ref{tab:prob_desc}. Table \ref{tab:gen_desc} lists all generator descriptions. Table \ref{tab:prob_desc} presents the problem descriptions of all COP settings. For demonstration purposes, we only present seed generators and design prompts for CVRP generator (prompt 7 and 9) and TSP S3 generator (prompt 8 and 10). The full ensemble of prompts are in our supplementary materials. 
 \end{minipage}

 \smallskip

 \vfill

 \begin{center}
 \begin{minipage}{\textwidth}
     \centering
     \includegraphics[width=0.94\textwidth]{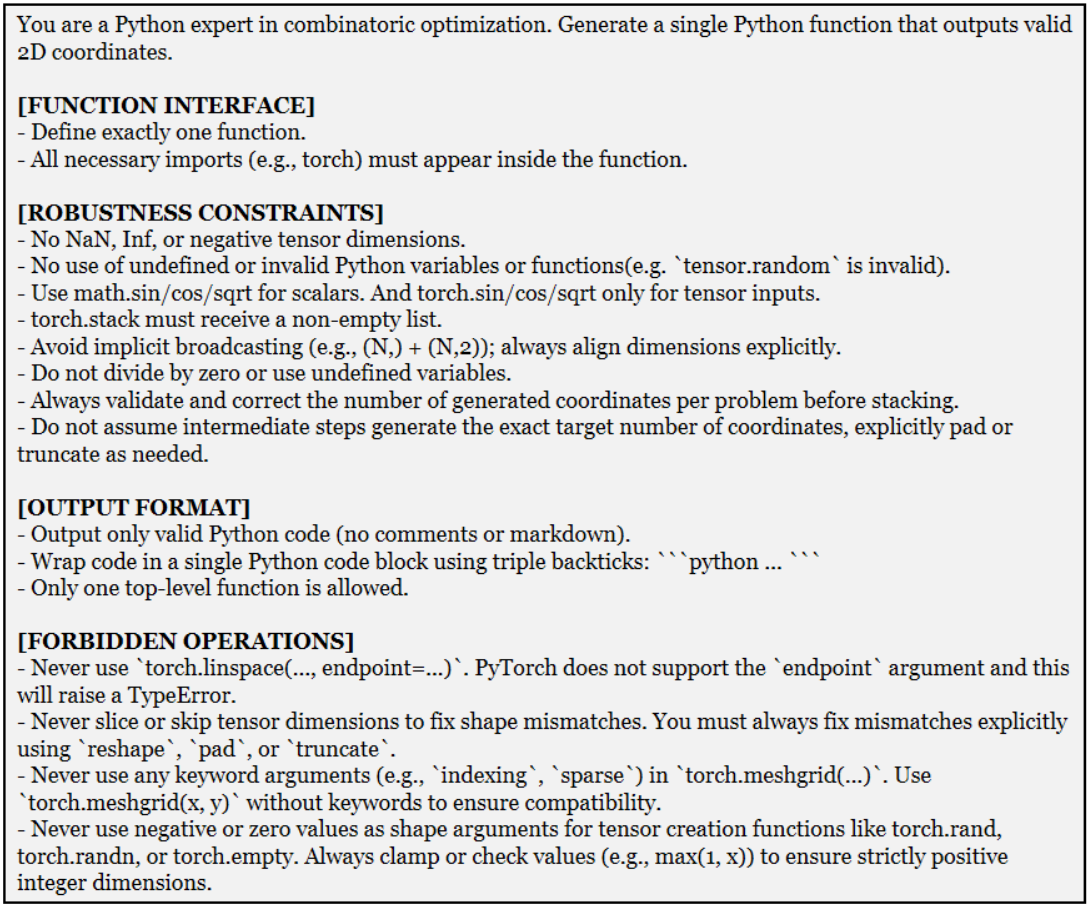}\\[3pt]
     Prompt 1: System prompt for generator LLM.\\[1em]
 \end{minipage}
 \end{center}

 \begin{center}
 \begin{minipage}{\textwidth}
     \centering
     \includegraphics[width=0.94\textwidth]{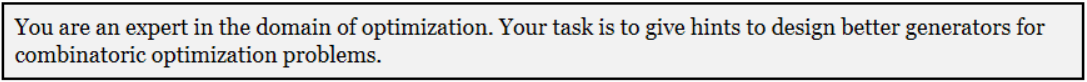}\\[3pt]
     Prompt 2: System prompt for reflector LLM.\\[1em]
 \end{minipage}
 \end{center}

 \begin{figure*}[t]
     \centering
     \includegraphics[width=0.94\textwidth]{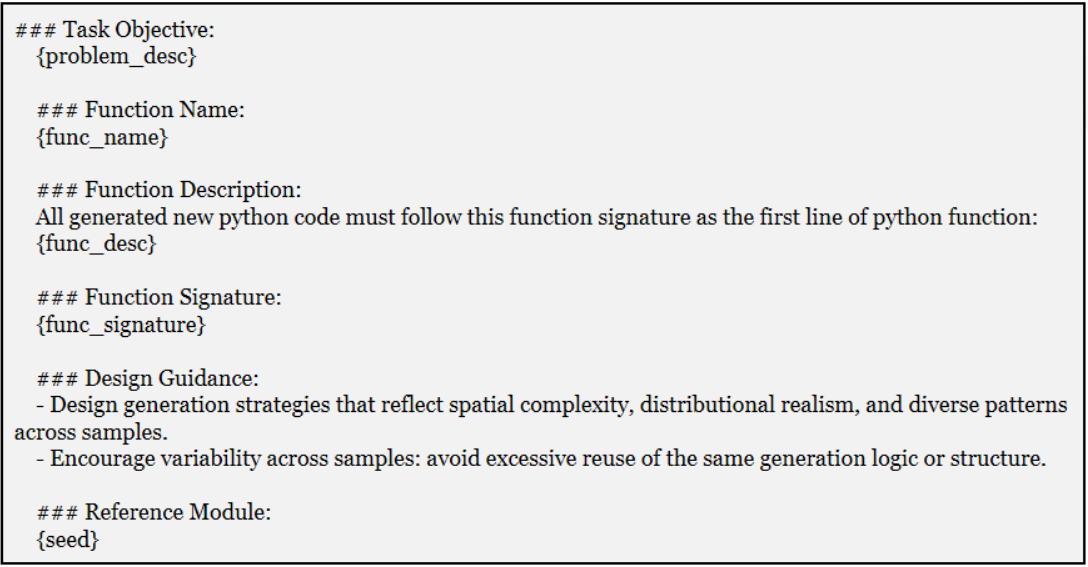}
     \caption*{Prompt 3: User prompt for generator population initialization.}
 \end{figure*}

 System prompts (prompt 1 and 2) are used to formalize and constrain the task and behavior-including generation procedures, standard formats, encouraged behaviors, and forbidden operations in performing the required task-for both the LLM generator and the LLM reflector. The user prompt (prompt 3) is used to construct the task-specific instructions, covering problem description, function name and signature, function description and seed prompt for EvoReal. The function signatures (prompt 4) here specify function names with their types. Seed prompts (prompt 5 and prompt 6) are designed to include the hand-crafted design prompt as well as the seed generator, which together specify the generator styles while narrowing the search space. The design prompts establish verbal boundaries that constrain the generator’s design style.

 \begin{figure*}[t]
     \centering
     \includegraphics[width=0.94\textwidth]{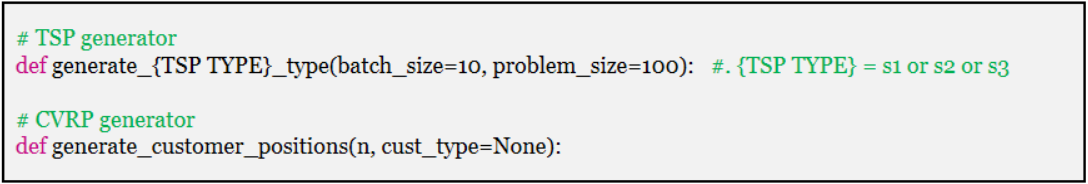}
     \caption*{Prompt 4: Function signatures used in EvoReal.}
 \end{figure*}

 \begin{figure*}[t]
     \centering
     \includegraphics[width=0.94\textwidth]{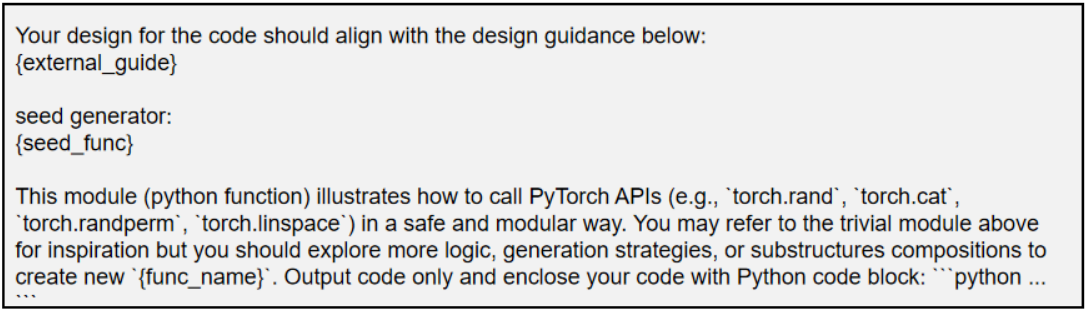}
     \caption*{Prompt 5: Seed prompt for TSP generator population initialization.}
 \end{figure*}

 \begin{figure*}[t]
     \centering
     \includegraphics[width=0.94\textwidth]{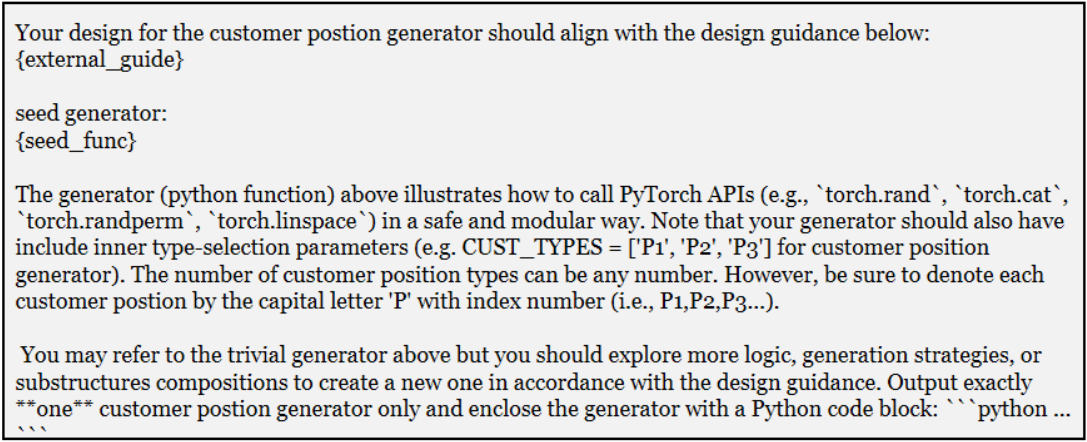}
     \caption*{Prompt 6: Seed prompt for CVRP generator population initialization.}
 \end{figure*}

 \begin{figure*}[t]
     \centering
     \includegraphics[width=0.94\textwidth]{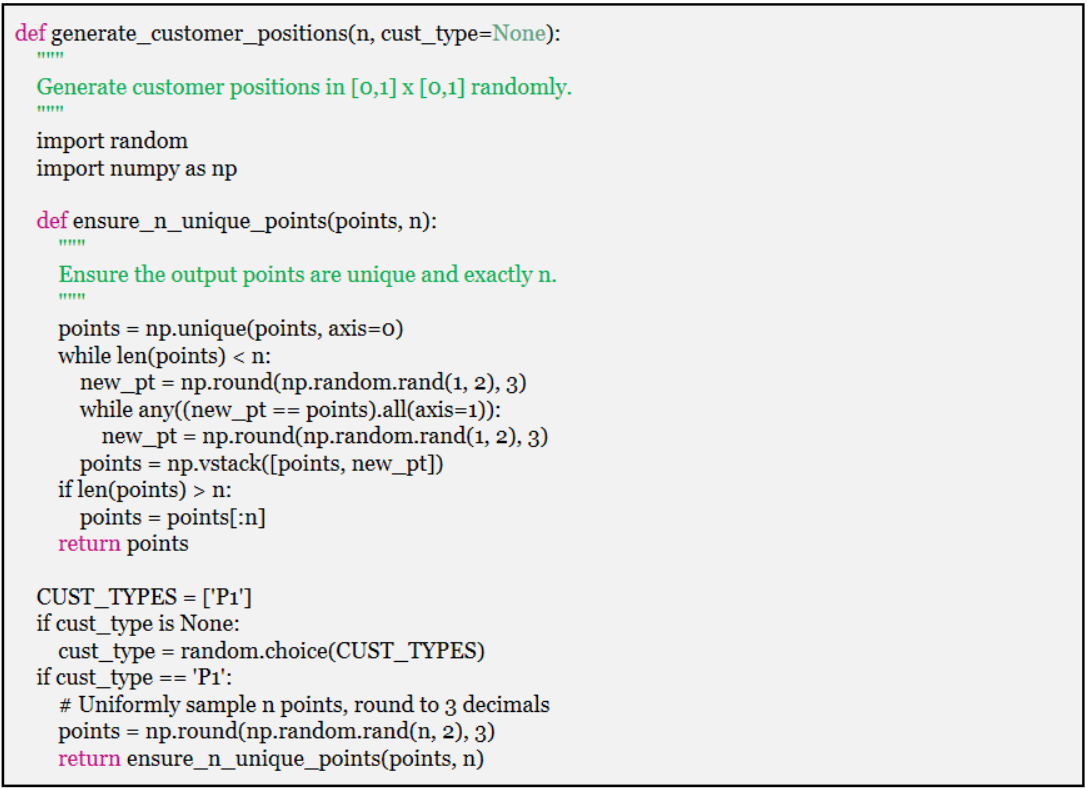}
     \caption*{Prompt 7: Seed CVRP coordinate generator used in EvoReal.}
 \end{figure*}

 \begin{figure*}[t]
     \centering
     \includegraphics[width=0.94\textwidth]{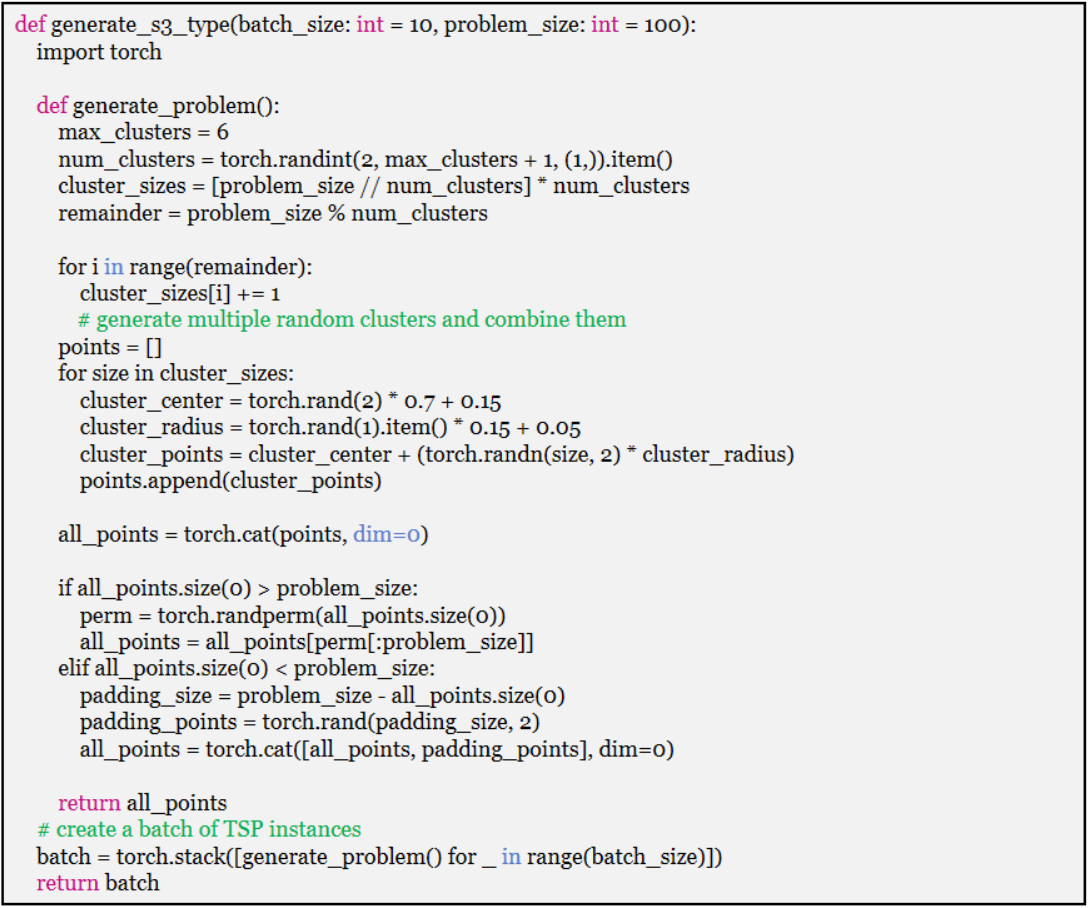}
     \caption*{Prompt 8: Seed TSP coordinate generator used in EvoReal.}
 \end{figure*}

 \begin{figure*}[t]
     \centering
     \includegraphics[width=0.94\textwidth]{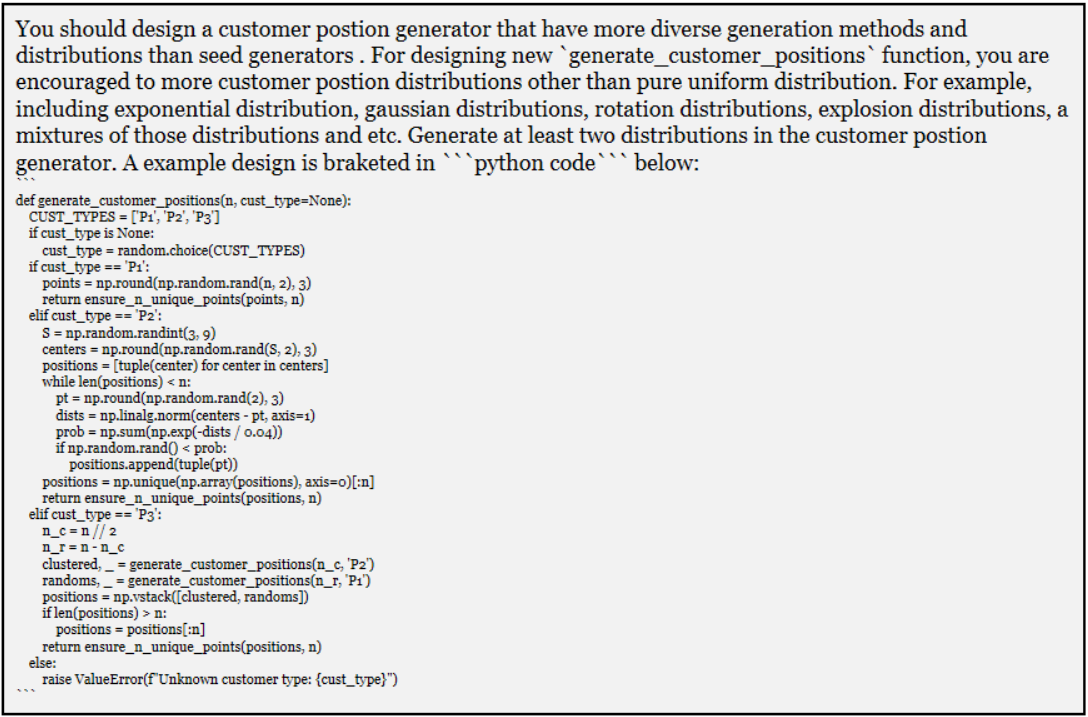}
     \caption*{Prompt 9: Design prompt (external knowledge) for CVRP coordinate generator.}
 \end{figure*}

 \begin{figure*}[t]
     \centering
     \includegraphics[width=0.94\textwidth]{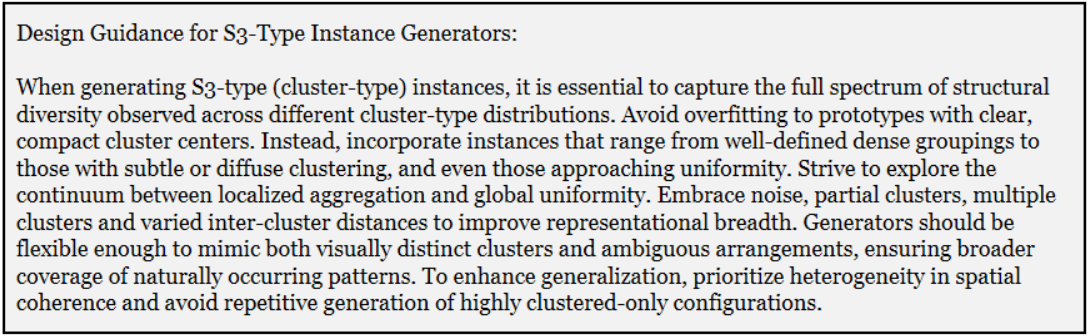}
     \caption*{Prompt 10: Design prompt (external knowledge) for TSP coordinate generator.}
 \end{figure*}

 \begin{table*}[t]
 \centering

 \begin{tabular}{p{0.18\textwidth}|p{0.77\textwidth}}
 \hline
 \textbf{Problem} & \textbf{Function description} \\
 \hline
 TSP\_generator &
 The function must take two arguments: \texttt{(batch\_size, problem\_size)}. It must return a batch of synthetic TSP problems as a \texttt{torch.Tensor} of shape \texttt{(batch\_size, problem\_size, 2)}, with \texttt{dtype=torch.float32}. Each individual problem must be well-structured and exactly of size \texttt{(problem\_size, 2)} to ensure safe stacking. The parameter \texttt{problem\_size} defines the number of nodes in each generated TSP instance. 
 All generated problems for stacking in the final return must have exactly \texttt{problem\_size} 2D coordinates. Use \texttt{torch.randperm} to truncate or \texttt{torch.cat + torch.rand()} to pad if needed. 
 The output must be stackable via \texttt{torch.stack} without shape errors. \\
 \hline
 CVRP\_generator &
 The function must take two arguments: \texttt{(n, cust\_type=None)}. It generates a 2D coordinate array of shape \texttt{(n, 2)} representing customer positions for a CVRP instance. The parameter \texttt{n} is the number of customers in each generated instance, while \texttt{cust\_type=None} is a fixed trivial parameter (do not change it). 
 The function must always return a \texttt{numpy.ndarray} of shape \texttt{(n, 2)}, where all coordinate values are within the range [0, 1], and all customer positions are unique. 
 To ensure uniqueness, the function must include and call the helper function below (do not modify or remove it):
 \begin{quote}
 \begin{verbatim}
 def ensure_n_unique_points(points, n):
     """Ensure the output points are unique and exactly n."""
     points = np.unique(points, axis=0)
     while len(points) < n:
         new_pt = np.round(np.random.rand(1, 2), 3)
         while any((new_pt == points).all(axis=1)):
             new_pt = np.round(np.random.rand(1, 2), 3)
         points = np.vstack([points, new_pt])
     if len(points) > n:
         points = points[:n]
     return points
 \end{verbatim}
 \end{quote}
 Both the input and the return of \texttt{ensure\_n\_unique\_points} are \texttt{(n, 2)} \texttt{numpy.ndarray}. \\
 \hline
 \end{tabular}
 \caption{Generator descriptions used in prompts.}
 \label{tab:gen_desc}
 \end{table*}

 \begin{table*}[t]
 \centering

 \begin{tabular}{p{0.18\textwidth}|p{0.77\textwidth}}
 \hline
 \textbf{Problem} & \textbf{Problem description} \\
 \hline
 TSP\_generator &
 The goal is to design a TSP problem data generator --- a Python function that outputs synthetic TSP instances for fine-tuning a neural solver. The generated problems should not merely be valid or diverse, but should exhibit structural patterns that are highly representative of a specific class of TSPLib distributions. This includes capturing repeated motifs, geometric arrangements, directional or symmetric alignments, and other spatial regularities. The generator should produce well-formed instances of fixed problem size and exhibit strong generalization on real TSPLib problems of the same type, serving as an abstract yet effective summary of the target distribution's geometric and distributional characteristics. \\
 \hline
 CVRP\_generator &
 The goal is to design a CVRP customer position generator --- a Python function that outputs synthetic CVRP customer coordinate positions for fine-tuning a neural solver. The generated problems should not merely be valid or diverse, but should exhibit structural patterns that are highly representative of a specific class of CVRPLib distributions. The generator should produce a well-formed set of customer coordinates of fixed problem size \texttt{n} and exhibit strong generalization on real CVRPLib problems of the same type, serving as an abstract yet effective summary of the target distribution's geometric and distributional characteristics. \\
 \hline
 \end{tabular}
 \caption{Problem descriptions used in prompts.}
 \label{tab:prob_desc}
 \end{table*}

\end{document}